\documentclass{article}
\usepackage{ijcai23}

\usepackage{times}
\usepackage{soul}
\usepackage{url}
\usepackage[hidelinks]{hyperref}
\usepackage[utf8]{inputenc}
\usepackage[small]{caption}
\usepackage{graphicx}
\usepackage{amsmath}
\usepackage{amsthm}
\usepackage{booktabs}
\usepackage{algorithm}
\usepackage{algorithmic}
\usepackage{amssymb}
\usepackage{subcaption}
\usepackage[switch]{lineno}

\usepackage{xcolor,colortbl}
\hypersetup{
	colorlinks   = true, 
	urlcolor     = red, 
	linkcolor    = red, 
	citecolor   = green 
}
\graphicspath{{images/}}
\definecolor{ceiling}{RGB}{214,  38, 40}   %
\definecolor{floor}{RGB}{43, 160, 4}     %
\definecolor{wall}{RGB}{158, 216, 229}  %
\definecolor{window}{RGB}{114, 158, 206}  %
\definecolor{chair}{RGB}{204, 204, 91}   %
\definecolor{bed}{RGB}{255, 186, 119}  %
\definecolor{sofa}{RGB}{147, 102, 188}  %
\definecolor{table}{RGB}{30, 119, 181}   %
\definecolor{tvs}{RGB}{160, 188, 33}   %
\definecolor{furniture}{RGB}{255, 127, 12}  %
\definecolor{objects}{RGB}{196, 175, 214} %

\definecolor{car}{rgb}{0.39215686, 0.58823529, 0.96078431}
\definecolor{bicycle}{rgb}{0.39215686, 0.90196078, 0.96078431}
\definecolor{motorcycle}{rgb}{0.11764706, 0.23529412, 0.58823529}
\definecolor{truck}{rgb}{0.31372549, 0.11764706, 0.70588235}
\definecolor{other-vehicle}{rgb}{0.39215686, 0.31372549, 0.98039216}
\definecolor{person}{rgb}{1.        , 0.11764706, 0.11764706}
\definecolor{bicyclist}{rgb}{1.        , 0.15686275, 0.78431373}
\definecolor{motorcyclist}{rgb}{0.58823529, 0.11764706, 0.35294118}
\definecolor{road}{rgb}{1.        , 0.        , 1.        }
\definecolor{parking}{rgb}{1.        , 0.58823529, 1.        }
\definecolor{sidewalk}{rgb}{0.29411765, 0.        , 0.29411765}
\definecolor{other-ground}{rgb}{0.68627451, 0.        , 0.29411765}
\definecolor{building}{rgb}{1.        , 0.78431373, 0.        }
\definecolor{fence}{rgb}{1.        , 0.47058824, 0.19607843}
\definecolor{vegetation}{rgb}{0.        , 0.68627451, 0.        }
\definecolor{trunk}{rgb}{0.52941176, 0.23529412, 0.        }
\definecolor{terrain}{rgb}{0.58823529, 0.94117647, 0.31372549}
\definecolor{pole}{rgb}{1.        , 0.94117647, 0.58823529}
\definecolor{traffic-sign}{rgb}{1.        , 0.        , 0.    }

\title{OccDepth: A Depth-Aware Method for 3D Semantic Scene Completion}

\author{
	Ruihang Miao\footnotemark[1]
	\and
	Weizhou Liu\footnotemark[1]\and
	Mingrui Chen\and
	Zheng Gong \and
	Weixin Xu  \and
	Chen Hu\footnotemark[2] \and
	Shuchang Zhou
	\affiliations
	MEGVII Technology
}

\begin{document}
	
	\maketitle
	\renewcommand{\thefootnote}{\fnsymbol{footnote}} 
	\footnotetext[1]{Equal contribution.} 
	\footnotetext[2]{Corresponding authors.} 

	\begin{abstract}
    3D Semantic Scene Completion (SSC) can provide dense geometric and semantic scene representations, which can be applied in the field of autonomous driving and robotic systems. It is challenging to estimate the complete geometry and semantics of a scene solely from visual images, and accurate depth information is crucial for restoring 3D geometry. In this paper, we propose the first stereo SSC method named OccDepth, which fully exploits implicit depth information from stereo images (or RGBD images) to help the recovery of 3D geometric structures. The Stereo Soft Feature Assignment (Stereo-SFA) module is proposed to better fuse 3D depth-aware features by implicitly learning the correlation between stereo images. In particular, when the input are RGBD image, a virtual stereo images can be generated through original RGB image and depth map. Besides, the Occupancy Aware Depth (OAD) module is used to obtain geometry-aware 3D features by knowledge distillation using pre-trained depth models. In addition, a reformed TartanAir benchmark, named SemanticTartanAir, is provided in this paper for further testing our OccDepth method on SSC task.
	Compared with the state-of-the-art RGB-inferred SSC method, extensive experiments on SemanticKITTI show that our OccDepth method achieves superior performance with improving +4.82\% mIoU, of which +2.49\% mIoU comes from stereo images and +2.33\% mIoU comes from our proposed depth-aware method. Our code and trained models are available at \url{https://github.com/megvii-research/OccDepth}.
	\end{abstract}
 
 	\section{Introduction}
		\begin{figure}[!t]
		\centering
		\includegraphics[width=0.45\textwidth]{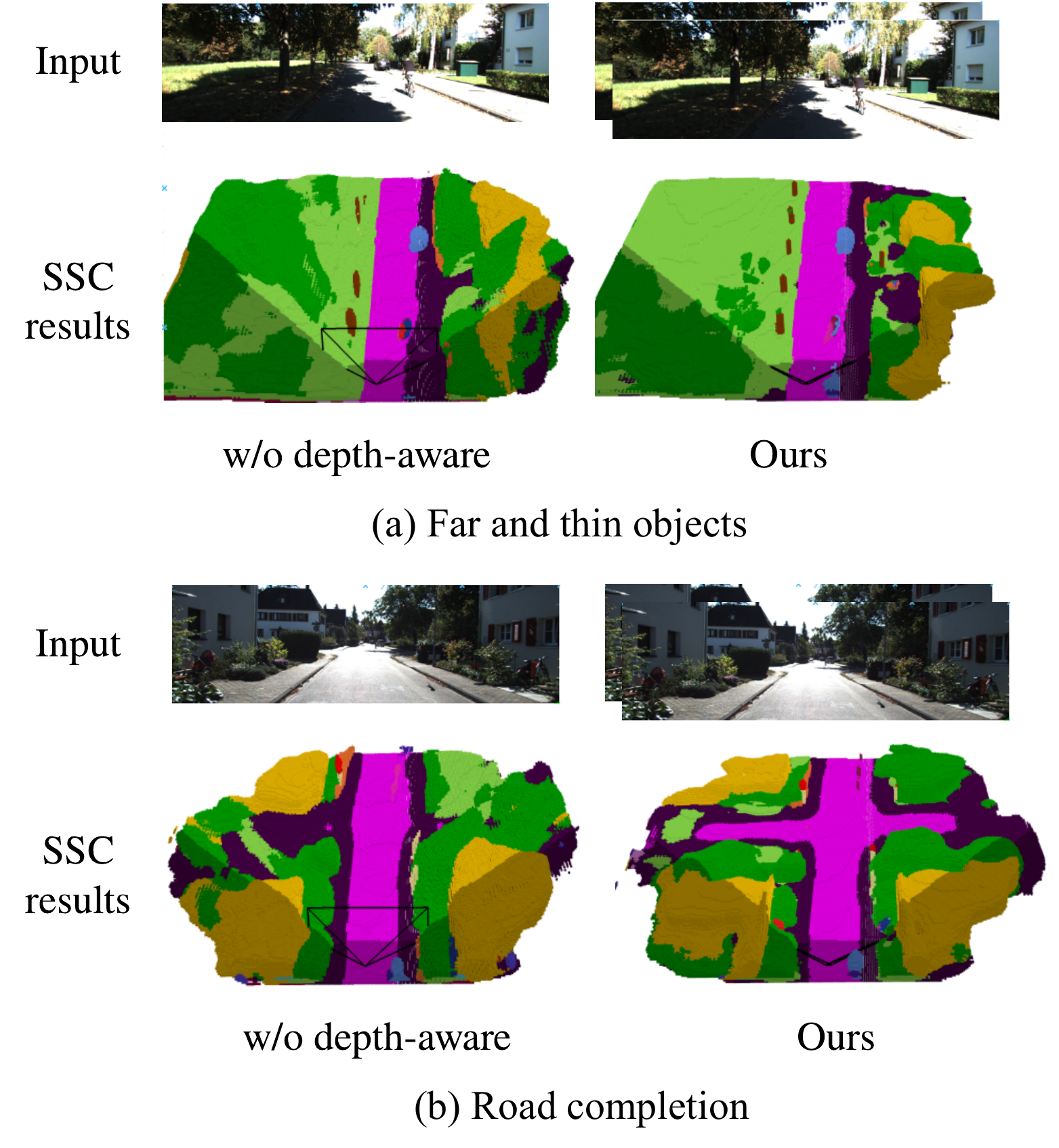}
		\caption{RGB based Semantic Scene Completion with/without depth-aware. (a) Our proposed OccDepth method can detect smaller and farther objects. (b) Our proposed OccDepth method complete road better.}
		\label{introShow}
	\end{figure}
 
	Humans can use their eyes to construct geometric and semantic information in 3D scenes, which can effectively help humans interact with 3D worlds. Dense representations of 3D scenes such as occupancy, dense point clouds, SDF \cite{park2019deepsdf}, etc. are useful for agents to perform various high-level tasks in a refined manner. Recently, we have witnessed great progress on these tasks, including path planning and environment interaction tasks. Numerous schemes have begun to explore the use of pure vision to restore dense 3D geometric structures and semantic information. However, restoring 3D structures from 2D images remains a challenging problem. 
	
	The 3D Semantic Scene Completion (SSC) task \cite{roldao20223d} attempts to reconstruct geometric and semantic structures based on observations and to complete occluded regions based on priors. The majority of the currently used 3D SSC technique depends on the input of 3D signals with depth information, such as voxels, point clouds, depth maps, and TSDF, which require expensive sensors to obtain. It is desired to explore schemes that are based on cheaper and easy-to-deploy image sensors. 
    Accurate depth information is crucial for the reconstruction and prediction of 3D geometric structures, while pure vision solutions frequently lack it. In the field of 3D detection, the accuracy of the pure vision solution is already competitive with the lidar solution. After the depth information is added to the 3D detection, it can be further improved, such as BEVDepth \cite{li2022bevdepth}. Therefore, we aim to reduce the gap between vision-only and 3D input solutions by using depth information to enhance the performance of pure vision for SSC task.
    
	In this paper, we explore how to exploit implicit depth information in images to help 3D SSC recover reliable geometry. Therefore, we introduce OccDepth, which is the first stereo 3D SSC method with vision-only input. In order to effectively use depth information, a Stereo Soft Feature Assignment (Stereo-SFA) module is proposed to better fuse 3D depth-aware features by implicitly learning the correlation between stereo images. An Occupancy Aware Depth (OAD) module uses knowledge distillation explicitly with pre-trained depth models to produce depth-aware 3D features. Additionally, we propose the SemanticTartanAir dataset based on TartanAir in order to validate the stereo-input scene more effectively. This is a virtual scene where real ground truth can be used to better evaluate the effectiveness of the method. Experiments show that our OccDepth method achieves superior performance on both SemanticKitti, NYUv2, and SemanticTartanAir benchmarks. Figure \ref{introShow} shows the visualized results of OccDepth. With only images input, OccDepth can reconstruct thin objects, far objects, and objects with distinct geometric structures better.
	In summary, the main contributions are as follows:
	\begin{itemize}
		\item A Stereo Soft Feature Assignment (Stereo-SFA) module that better fuses 3D depth-aware features by implicitly learning correlation between stereo images.
		\item An Occupancy Aware Depth (OAD) module that uses knowledge distillation explicitly with pre-trained depth models to produce depth-aware 3D features.
		\item Our experiments show that our OccDepth method outperforms all vision-only baseline methods by a large margin and is close to the 3D input solutions.
	\end{itemize}
	\section{Related works}
	\textbf{3D Reconstruction From Images} is a long-standing problem. Early methods use shape-from-shading \cite{durou2008numerical} or structure-from-motion \cite{schonberger2016structure,cui2017hsfm}. Recent neural radiance fields (NeRF) based methods \cite{mildenhall2021nerf} learn the density and color fields of 3D scenes. Built on the generalizable SceneRF \cite{cao2022scenerf} introduces a probabilistic ray sampling strategy and a spherical U-Net, to reconstruct large scenes using only single view as input. In addition to the above 3D reconstruction task, Scene Completion (SC) infers the complete geometry of a scene given images and/or 3D sparse inputs. SG-NN \cite{dai2020sg} converts partial and noisy RGB-D scans into high-quality 3D scenes. Recently, \cite{engelmann2021points,li2021joint,liu2021voxel,sharma2022seeing} start to explore the 3D Semantic Instance Completion task, in which the object instances and their reconstructions are predicted from partial observations of the scene. In our work, we focus on the completion of geometry and semantics, named 3D SSC, that provides useful information for high-level tasks. 
	
	\textbf{3D SSC} is introduced in SSCNet \cite{song2017semantic}, which shows that learning semantic segmentation and scene completion can mutually benefit. SSCNet takes a single RGB-D image as input and outputs a 3D voxel representation of occupancy and semantic labels. Following SSCNet, \cite{garbade2019two,li2020attention} leverage the depth and semantic information to achieve a better SSC result. Moreover, \cite{tang2022not} transfers voxels to point clouds to reduce the computation redundancy, and then fuses the voxel and the point stream using the proposed anisotropic voxel aggregation operator. In addition to the RGB-D based SSC methods, works \cite{cheng2021s3cnet,wu2020scfusion,yang2021semantic,yan2021sparse,rist2021semantic} based on geometrical inputs (depth, point cloud or occupancy grids) are also proposed. Recently, \cite{dahnert2021panoptic}, an only RGB image based method, unifies the tasks of geometric reconstruction, 3D semantic segmentation, and 3D instance segmentation of indoor scenes from a single RGB image. MonoScene \cite{Cao_2022_CVPR} infers dense 3D voxelized semantic indoor and outdoor scenes from a single RGB image. Similar to MonoScene, we use multiview RGB images to estimate dense geometry and semantics for indoor and outdoor scenes.

	\textbf{Knowledge Distillation} is initially proposed in \cite{hinton2015distilling}. The main idea is to transfer the learned knowledge from a teacher network to a student one. Such a transference can be achieved by the soft targets of the output layer \cite{hinton2015distilling}, the intermediate feature map \cite{romero2014fitnets} and the activation status of each neuron \cite{huang2017like}. Knowledge distillation has been widely investigated in a variety of computer vision tasks, such as semantic segmentation \cite{hou2020inter,liu2019structured}, 3D object detection \cite{chong2022monodistill,zheng2022boosting} and depth prediction \cite{pilzer2019refine,ye2019student}. Among the depth prediction methods, BEVDepth \cite{li2022bevdepth} leverages explicit depth supervision through the sparse depth information from LiDAR to improve the accuracy of depth prediction. Different from BEVDepth, we utilize the explicit dense multiview stereo depth to teach the OAD module learn the depth prediction capability, which is more effective for stereo image inputs.
	
	\section{Methods}
	\subsection{Overview}
	\begin{figure*}[!t]
		\centering
		\includegraphics[width=0.97\textwidth]{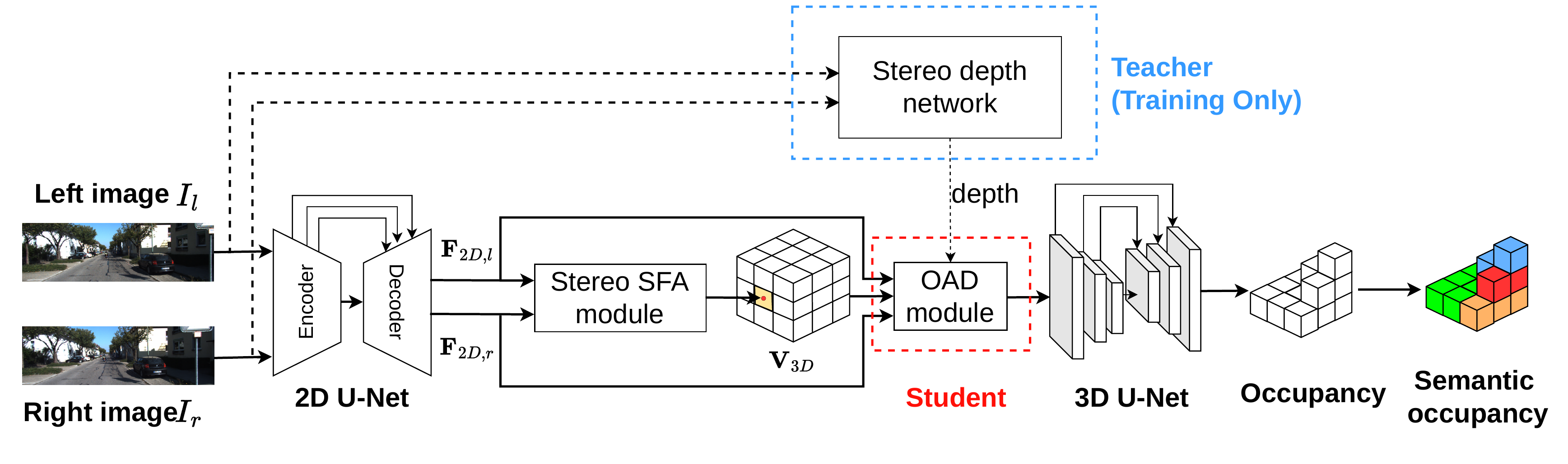}
		\caption{The process pipeline of the proposed OccDepth. The 3D SSC is inferred from stereo images with bridging a Stereo-SFA module to lift features to 3D space, an OAD module to enhance depth prediction, and a 3D U-Net to extract geometry and semantics. The stereo depth network is only used in training for giving a depth supervision.}
		\label{processPipeline}
	\end{figure*}
	The overall processing pipeline of our OccDepth network is shown in Figure \ref{processPipeline}. Stereo images $I_{l}$, $I_{r}$ are taken as inputs and are encoded into 2D features $\mathbf{F}_{2D,l}, \mathbf{F}_{2D,r}\in\mathbb{R}^{H\times W\times C}$. Then the 2D features are fused into 3D voxels, where the implicit depth information is embedded through a stereo soft feature assignment (Stereo-SFA) module. Next the explicit depth information is added into 3D features through an occupancy aware depth (OAD) module. Subsequently, during calculating the loss, the overall semantic scene completion loss are decoupled into geometric loss and semantic loss for better prediction. At last, a voxelized 3D scene $\hat{y}$ with class labels $\mathcal{C}=\{c_0,c_1,...,c_N\}$ is predicted, where $N$ is the total number of the semantic classes. As "empty" can also be regarded as a semantic class, $c_0$ represents empty voxels.
	
	The rest of this section will describe the Stereo-SFA module (Sec. \ref{sectionFeaturePrjection}), OAD module (Sec. \ref{sectionOccupancyAwareDepth}), losses (Sec. \ref{sectionGeometricSemanticDecouplingModule}) and tricks for mitigating over-fitting (Sec. \ref{sectionTMO}).
	
	\subsection{Stereo Soft Feature Assignment Module}
	\label{sectionFeaturePrjection}
	\begin{figure}[!t]
		\centering
		\includegraphics[width=0.49\textwidth]{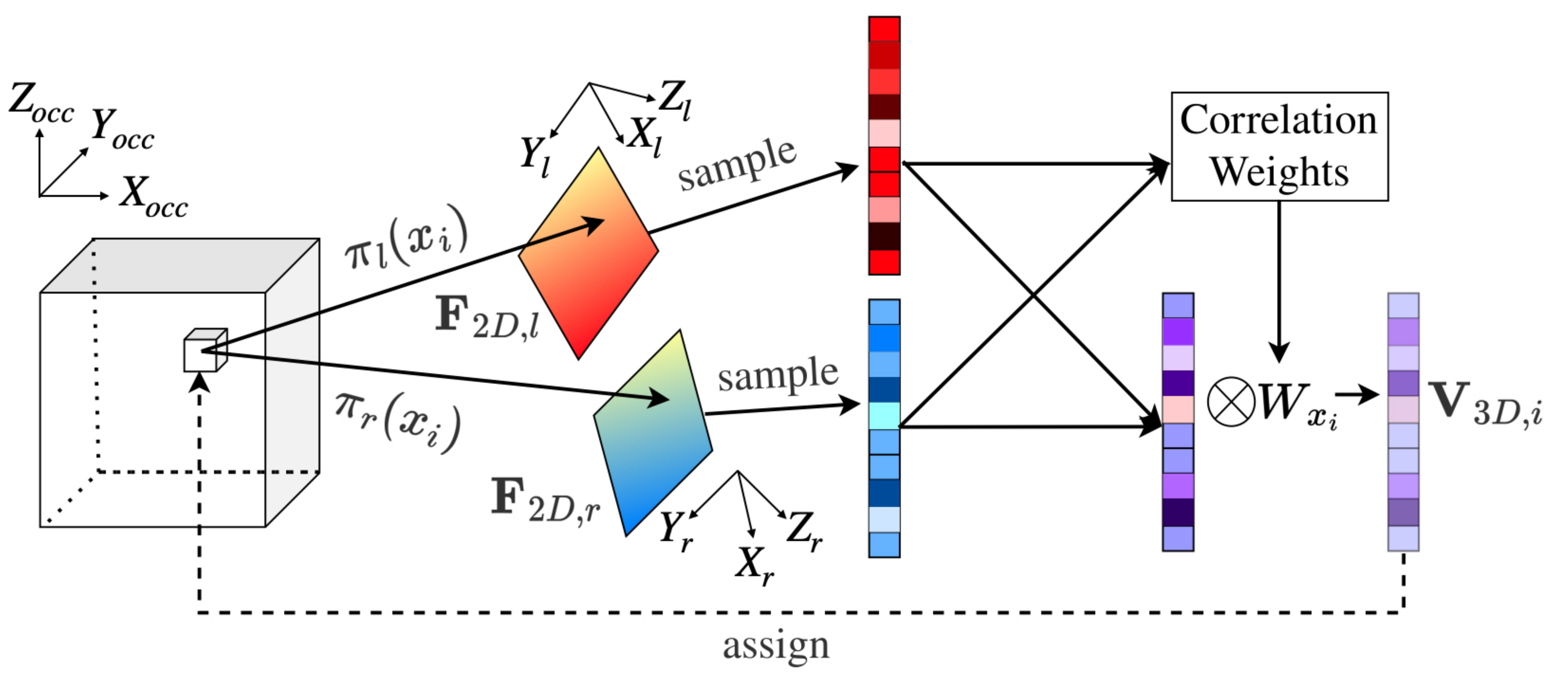}
		\caption{The illustration of stereo soft feature assignment module. The sampled 2D features are fused to 3D voxel feature.}
		\label{StereoFeatureAssignment}
	\end{figure}
	In semantic occupancy predicting task, designing a good 2D-3D lifting function is a critical problem, especially when there is no direct depth information. While taking monocular image as input, the scale ambiguity is hard to handle when lifting 2D to 3D \cite{fahim2021single}. Different from monocular image, the stereo image can provide implicit depth information that solves the scale ambiguity to a certain extent.
	
	In order to calculate representation features of each voxel in the bounded 3D scene, the feature mapping between 2D image features and voxel features are needed. Unlike LSS \cite{philion2020lift} which implicitly unprojects 2D features to 3D space, we choose to project each voxel to the corresponding image pixel as done in MonoScene \cite{Cao_2022_CVPR}, which establishes a full feature mapping for all voxels. Moreover, the implicit depth information, which is derived through calculating the correlation between the left and right 2D features, is encoded as the weights of 3D features. Ablation in Sec. \ref{AblationStudy} shows that the above proposed feature mapping method is better than directly using stereo depth and is better than using a combination of left 2D features and right 2D features. 
	
	Figure \ref{StereoFeatureAssignment} shows the process of Stereo-SFA module. Assuming that the stereo cameras has been calibrated and the input images have been undistorted, the intrinsic and extrinsic matrices of stereo cameras are known. Given $X\times Y\times Z$ voxels with centroid at coordinate $\mathbf{x} \in \mathbb{R}^{X\times Y\times Z\times 3}$, the 3D-2D projecting can be denoted as $\pi(\mathbf{x})$. Then, the 3D voxel feature $\mathbf{V}_{3D} \in \mathbb{R}^{X\times Y\times Z\times C}$ can be sampled from corresponding 2D feature map $\mathbf{F}_{2D}$ and written as
	\begin{equation}
	\mathbf{V}_{3D}=\phi_{\pi(\mathbf{x})}(\mathbf{F}_{2D}),
	\end{equation}
	where $\phi_\mathbf{x}(M)$ is the sampling function that sample feature mapping $M$ at coordinates $\mathbf{x}$. The 3D features will be set as $\mathbf{0}$ when the projected 2D points are out of the image. 
	
	Denoting $\mathbf{V}_{3D,l}$ and $\mathbf{V}_{3D,r}$ as the 3D features acquired from the left 2D feature map and the right 2d feature map respectively, the weighted 3D feature $\mathbf{V}_{3D,w}$ is written as
	\begin{equation}
	\mathbf{V}_{3D,w} =\begin{cases}
	\mathbf{V}_{3D,l},&\text{if $\mathbf{V}_{3D,r}=\mathbf{0}$}\\
	\mathbf{V}_{3D,r},&\text{if $\mathbf{V}_{3D,l}=\mathbf{0}$}\\
	w\times \frac{\mathbf{V}_{3D,l}+\mathbf{V}_{3D,r}}{2},&\text{others}\
	\end{cases},
	\end{equation}
	where $w$ denotes the weights calculated from the correlation between $\mathbf{V}_{3D,l}$ and $\mathbf{V}_{3D,r}$. In this paper, the cosine similarity is used to measure the correlation of features.
	In order to enlarge the receptive field, the multi-scale 2D feature maps are considered. The output 3D feature map is written as
	\begin{equation}
	\mathbf{V}_{3D}=\sum_{s\in \mathcal{S}}{\mathbf{V}^s_{3D,w}},
	\end{equation}
	where $\mathcal{S}=\{1,2,4,8\}$ is a set of downsampling scales.

	\subsection{Occupancy-Aware Depth Module}
	\label{sectionOccupancyAwareDepth}
	\subsubsection{Module Structure}
	\begin{figure}[!t]
            \flushright
		\includegraphics[width=0.49\textwidth]{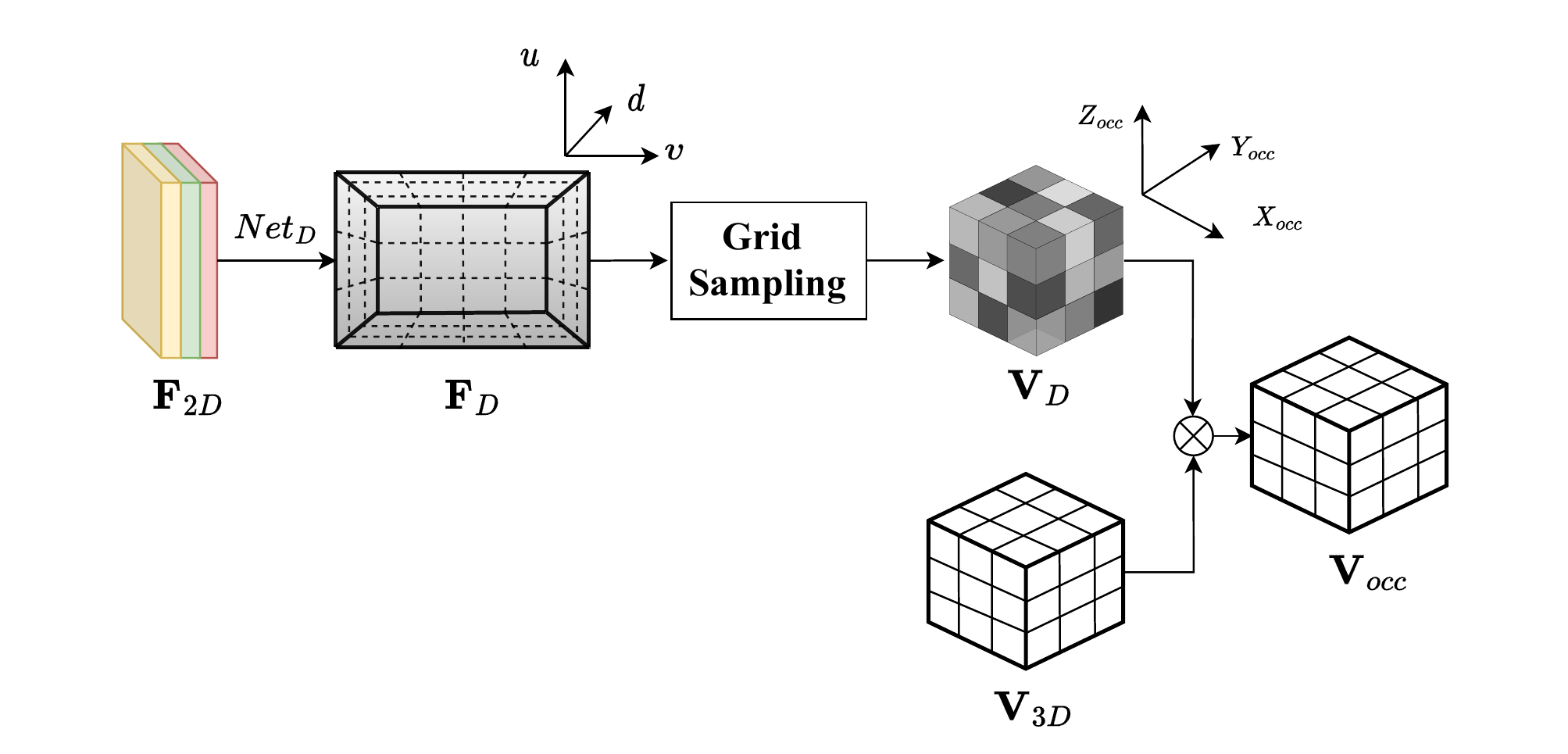}
		\caption{The illustration of the occupancy aware depth module. For simplicity, only the processing flow of single-shot $V_{D}$ is shown. The OAD module is used to introduce spatial occupancy prior by the predicted depth information.}
		\label{sectionOccupancyAwareDepth}
	\end{figure}
	In order to introduce spatial occupancy prior information when converting 2D image feature to 3D voxel feature, the OAD module is proposed.
	Inspired by the excellent 3D object detection works \cite{li2022unifying,reading2021categorical}, the OAD module uses the predicted depth information to estimate the prior probability of the existence of objects in the voxel feature space. This probability information is then used to improve the spatial information of voxel features.
	
	Figure \ref{sectionOccupancyAwareDepth} shows the process of our occupancy aware depth module.
	Through the above Stereo-SFA module, the corresponding stereo feature maps in the voxel space $\mathbf{V}_{3D} \in \mathbb{R}^{X\times Y\times Z\times C}$ are obtained.
	The single scale image feature $\mathbf{F}_{2D} \in \mathbb{R}^{H \times W \times C}$ with downsampling scale $S$ is fed to the OAD module, where $S$ is set to be 8.
	Firstly, a depth distribution network $Net_{D}$ is used to predict the depth feature $\mathbf{F}_{D} \in \mathbb{R}^{C \times H\times W\times D}$ of the multi-view input images.
	Secondly, a softmax operator is used to transform the $\mathbf{F}_{D}$ into frustum depth distribution $\mathbf{G}_{D} \in \mathbb{R}^{H\times W\times D}$, where the categories are the $D$ discretized depth bins.
	Thirdly, the frustum depth distribution $\mathbf{G}_{D}$ is transformed to a voxel-space depth distribution representation $\mathbf{V}_{D} \in \mathbb{R}^{X\times Y\times Z}$ with the differentiable grid sampling process, using the camera calibration matrix $P \in \mathbb{R}^{3\times 4}$.
	There are several depth discretization methods: uniform discretization (UD), linear-increasing discretization (LID) \cite{tang2021center3d} and spacing-increasing discretization (SID) \cite{fu2018deep}.
	We utilize LID in depth discretization process as it provides a more refined expression for the near space. LID is defined as:
	\begin{equation}
	d_{c} = d_{min} + \frac{d_{max} - d_{min}}{D \cdot (D + 1)} \cdot d_i \cdot (d_i + 1),
	\end{equation}
	where $d_{c}$ and $d_{i}$ are continuous depth value and discretized depth bin index, [$d_{min}$, $d_{max}$] is the full depth range, and $D$ is the number of target discretized bins.
	Finally, the occupancy aware voxel feature $\mathbf{V}_{occ}$ can be obtained:
	\begin{equation}
	\mathbf{V}_{occ} = (\sum_{i}^{2}{\mathbf{V}_{D}^{i} \cdot \mathbf{M}}) \cdot \mathbf{V}_{3D},
	\end{equation}
	where $\mathbf{M}$ is the mask for averaging the voxel pixel in the overlap region between stereo inputs, the value in overlap region is 0.5 and others is 1.0, and $\mathbf{V}_{D}$ can be represented as occupancy probability prior in voxel space.
	
	\subsubsection{Stereo Depth Distillation}
	
	In the above mentioned module, the depth distribution network $Net_{D}$ is trained based on implicit supervision, where the gradient comes from the back propagation process of occupancy related loss.
	Because the 2D to 3D transformation process mentioned above is highly dependent on the performance of the predicted depth distribution.
	Inspired by BevDepth \cite{li2022bevdepth}, stereo depth distillation is adopted in the training process to help the depth distribution network $Net_{D}$ have better depth estimation performance.
	Stereo depth distillation introduces explicit supervision on the depth feature $\mathbf{F}_{D}$. Different from the BevDepth that uses the sparse lidar point cloud to generate ground truth depth, our ground truth depth $\mathbf{D}_{GT}$ with more dense information is generated by a stereo depth network LEAStereo.
	To align the ground truth depth with the predicted depth feature $\mathbf{F}_{D}$, a \textit{min pooling} and a \textit{one hot} operator are adopted on $\mathbf{D}_{GT}$ to generate $\mathbf{F}_{GT_{D}}$.
	Finally, binary cross entropy is adopted to calculate the depth loss:
	\begin{equation}
	L_{depth} = \psi(\mathbf{F}_{D}, \mathbf{F}_{GT_{D}}),
	\end{equation}
	where $\psi$ is the cross-entropy loss function.

	\subsection{Losses}
	\label{sectionGeometricSemanticDecouplingModule}
	As the semantic scene completion task can be decomposed into the occupancy prediction task and the semantic prediction task.
	Inspired by \cite{chen2020real}, we solve the geometric prediction task and the semantic prediction task step by step.
	Firstly, the $\mathbf{V}_{3D}$ generated from 3D U-Net structure can be processed to $\mathbf{F}_{3D}^{2class}$ with a single 3D convolutional layer,
	where the $\mathbf{F}_{3D}^{2class}$ is used to predict whether each element in the voxel is occupied.
	The $\mathbf{F}_{3D}^{2class}$ is used to generate occupancy loss:
	\begin{equation}
	L_{occ} = \psi(\mathbf{F}_{3D}^{2class}, \mathbf{GT}_{3D}^{geo}),
	\end{equation}
	where $\mathbf{GT}_{3D}^{geo}$ is the geometric ground truth with 2 classes, the first class represents the empty voxel and the second class represents the voxel that is occupied, and $\psi$ is the binary cross-entropy loss function.
	Secondly, the $\mathbf{F}_{3D}$ and $\mathbf{F}_{3D}^{2class}$ are concatenated as $\mathbf{F}_{3D}^{occ}$, and $\mathbf{F}_{3D}^{Nclass}$ is generated with a single 3d convolutional layer.
	The $\mathbf{F}_{3D}^{Nclass}$ is used to generate semantic loss:
	\begin{equation}
	L_{sem} = \psi(\mathbf{F}_{3D}^{Nclass}, \mathbf{GT}_{3D}^{sem}),
	\end{equation}
	where $\mathbf{GT}_{3D}^{sem}$ is the semantic ground truth with N classes corresponded to the dataset definition, and $\psi$ is the cross-entropy loss function.
	
	Refer to the definition of losses in Monoscene \cite{Cao_2022_CVPR}, $L_{mono}$ loss is also used in the 
 OccDepth training process. The $L_{mono}$ is defined as:
     \begin{equation}
	L_{mono} =  L_{rel} + L^{sem}_{scal} + L_{scal}^{geo} + L_{fp}.
     \end{equation}
      where the $L_{rel}$ is helpful for capturing better semantic information, the $L^{sem}_{scal}$  and $L_{scal}^{geo}$ are helpful for semantic and geometric accuracy improvement and the $L_{fp}$ is used to improve the completion ability. 
	So the OccDepth is trained by optimizing the following loss:
      \begin{equation}
	L_{total} = L_{occ} + L_{sem} + L_{depth} +  L_{mono}.
     \end{equation}
	\begin{table}[!t]
		\flushleft
		\begin{subtable}[!]{0.5\textwidth}
			\centering
   			\resizebox{0.9\textwidth}{!}{
				\begin{tabular}{l|c|c|c}
					\toprule
					&  &SC& SSC \\
					Method & Input &IoU & mIoU \\
					\midrule
     				\textbf{2.5D/3D}&&&\\
					LMSCNet$^{st}$&OCC & 33.00&5.80 \\
					AICNet$^{st}$&RGB, DEPTH & 32.8 &6.80 \\
					JS3CNet$^{st}$& PTS &39.30 &9.10 \\
                    \midrule
                    \textbf{2D}&&&\\
					MonoScene&RGB&34.16 & 11.08\\
					MonoScene$^{st}$&Stereo RGB&40.84 &13.57 \\
					OccDepth (ours)& Stereo RGB & \textbf{45.10}& \textbf{15.90}\\
					\bottomrule
			\end{tabular}}
			\caption{SemanticKITTI (hidden test set)}
			\label{mainResults0}
		\end{subtable}
		
		\begin{subtable}[!]{0.5\textwidth}
			\centering
   			\resizebox{0.9\textwidth}{!}{
				\begin{tabular}{l|c|c|c}
					\toprule
					&  &SC& SSC \\
					Method & Input & IoU& mIoU \\
					\midrule
                        \textbf{2.5D/3D}&&&\\
					LMSCNet$^{st}$&OCC &37.63&12.38 \\
					
					AICNet$^{st}$& RGB, DEPTH&31.86&10.03\\
					
					JS3CNet$^{st}$&PTS &26.81 & 9.55\\
                        3DSketch$^{st}$&RGB, TSDF &\textbf{62.46} &25.79\\
				\midrule
                    \textbf{2D}&&&\\	
                    MonoScene&RGB&34.87&24.12 \\
					MonoScene$^{st}$&Stereo RGB&49.63&32.79 \\
					OccDepth (ours)&Stereo RGB&52.49&\textbf{34.96}\\
					\bottomrule
			\end{tabular}}
			\caption{SemanticTartanAir (test set)}
			\label{mainResults2}
		\end{subtable}
		\caption{Performance on SemanticKITTI and SemanticTartanAir. The scene completion (SC IoU) and semantic scene completion (SSC mIoU) are reported for modified baselines (marked with superscript "st") and our OccDepth.}
		\label{mainResults}
	\end{table}	
	\subsection{Tricks for Mitigating Over-fitting}
	\label{sectionTMO}
	Due to the large amount of model parameters including 3D convolutions and the small amount of training data, the training process is prone to over-fitting problems.
	Several tricks are proposed to mitigate the over-fitting problem.  
	\begin{itemize}
		\item 2D Pre-training: The 2D backbone in the OccDepth can be pre-trained on a large instance segmentation datasets, the pre-training process can enhance semantic information in 2d features ($\mathbf{F}_{2D}$).
		\item Data Augmentation: Stronger data augmentation can mitigate the problem of having less training data. We find that Gaussian blur, grayscale and hue adjustment are helpful augmentation methods.
		\item Loss Weight Adjustment: Due to the imbalance of loss scales of different tasks, some over-fitting losses can greatly reduce the optimization effect of other losses. The over-fitting loss $L^{scal}_{sem}$ will be weighted by a gradually reduced weight $\gamma$ in the training process.The $\gamma$ is calculated as follows: $\gamma = max(0.2, 1-\frac{x}{N})$, where x is the current training step and N is the total training steps.
	\end{itemize}

 	\begin{table}[!t]
		\centering
            \resizebox{0.5\textwidth}{!}{
		\begin{tabular}{l|cc|cc|cc}
			\toprule
			 &\multicolumn{2}{c}{SemanticKITTI}& \multicolumn{2}{c}{SemanticTartanAir}& \multicolumn{2}{c}{NYUv2} \\
			Method  & IoU& mIoU& IoU& mIoU& IoU& mIoU \\
			\midrule
			\textbf{2.5D/3D}&&&&&&\\
			LMSCNet& \textbf{56.7}$^*$ & 17.6$^*$&92.2&36.5&44.1$^*$ & 20.4$^*$\\
			
			JS3CNet& 56.6$^*$ & 23.8$^*$&91.6&25.5&/ &/ \\
			
			S3CNet& 45.6$^*$ & \textbf{29.5}$^*$&/&/&/ &/\\
			
			Local-DIFs & \textbf{57.7}$^*$ &22.7$^*$&/&/& / &/ \\
			AICNet&/ &/&\textbf{99.0} &\textbf{40.0}& 43.9$^*$&23.8$^*$\\
                3DSketch&/ &/&91.6&31.5&49.5$^*$& 29.2$^*$\\
			
			\hline
			\textbf{2D}&&&&&&\\
			MonoScene & 34.2$^*$ & 11.1$^*$&34.9&24.1& 42.1 & 26.5\\
			
			OccDepth(ours) & 45.1 & 15.9&52.5&35.0&\textbf{50.3}  & \textbf{30.9}\\
			\bottomrule
		\end{tabular}}
		\caption{Baselines of 2.5D/3D-input methods. OccDepth outperforms some indoor baselines and is even comparable to some outdoor baselines. "$^*$" means results are cited from MonoScene. "/" means missing results.}
		\label{Baselines}
	\end{table}	
	
	\section{Experiments}
	OccDepth method is evaluated on SemanticKitti \cite{behley2019semantickitti}, NYUv2 \cite{silberman2012indoor} and SemanticTartanAir. The baseline methods are MonoScene \cite{Cao_2022_CVPR} and the stereo adaptation version of recent 2.5D/3D-input based SSC methods. we train our model for 30 epochs with an AdamW \cite{loshchilov2017decoupled} optimizer for all cases. The training hyper-parameters are: batchsize is 1, learning rate is $10^{-4}$, weight decay is $10^{-4}$. The learning rate is divided by 10 at epoch 20.
	
	In the rest of this section, dataset and metrics are described in Sec. \ref{DatasetsandMetric}. Then the comparison results between our method and SOTA methods are showed in Sec. \ref{ComparisonSOTA}. Finally, the ablation study about proposed modules are presented in Sec. \ref{AblationStudy}.
	\subsection{Datasets and Metric}
	\label{DatasetsandMetric}
	
	\paragraph{Datasets.} SemanticKITTI \cite{behley2019semantickitti} dataset provided $256 \times 256 \times 32$ voxel grids labeled with 21 classes. The voxels are generated through post-processing the Lidar scans and the size of each voxel is $0.2m$. We use cam2 and cam3 as a stereo RGB camera. The official 3834/815 train/val splits are used in all evaluations. The evaluating results of quantitative analysis are acquired from the hidden test set and the evaluating results of ablation studies are acquired from the validation set.
	
	NYUv2 \cite{silberman2012indoor} dataset provided $240\times 144\times 240$ voxels grids labeled with 13 classes. 795/654
	train/test splits are used in all evaluations as \cite{Cao_2022_CVPR,li2019rgbd}. The evaluation is taken on the downsampled test set with $60\times 36\times 60$ voxels. Because there is no stereo image on NYUv2, the Stereo-SFA module is removed in the experiments on NYUv2.
	
	SemanticTartanAir dataset based on TartanAir \cite{tartanair2020iros} is collected in a simulation environments that provides multimodal sensor data, including stereo RGB images, depth maps, segmentations, LiDAR point clouds, with precise ground truth labels. Although TartanAir dataset covers a wide range of scenes, we only focus on the indoor scene. Thus, we generate $120\times 48\times 120$ voxel grids with 14 classes of semantic labels for each RGB image in office scene with the help of the depth maps. The size of each voxel is $0.1m$. In SemanticTartanAir benchmark, 1976/192 train/test splits are used.
	
	\paragraph{Evaluation Metric.} Following previous works, the intersection over union (IoU) of occupied voxels and the mean IoU (mIoU) of semantic classes are reported for evaluating SC task and SSC task, respectively. In all datasets, the IoU and mIOU is calculated through on all voxels of the scene.

	\subsection{Comparison to State-of-the-arts}
	\label{ComparisonSOTA}
	We compare OccDepth with state-of-the-art SSC methods. These methods can be classified as RGB-inferred methods and 2.5D/3D-input methods.

 	\textbf{Comparison with stereo-based methods.} The performance of OccDepth on SemanticKITTI (hidden test set) and SemanticTartanAir (test set) is reported in Table \ref{mainResults}. All of the 2.5D/3D-input baseline methods are modified to stereo RGB-inferred version which are marked with superscript "st". Stereo images are used to generate occupancy, TSDF, point clouds and depth maps for 2.5D/3D-input baselines methods. For the monocular-based MonoScene \cite{Cao_2022_CVPR}, we fused 3D voxel features by average from stereo inputs, which named MonoScene$^{st}$. The best results are shown in bold. As there is no stereo images in NYUv2 dataset, we only compare  OccDepth with the original baselines in Table \ref{Baselines}.
	The experiment results demonstrate that our OccDepth outperforms other methods. Compared with MonoScene, which is a 2D baseline method, OccDepth improves about +4.82 mIoU/+10.94 IoU on SemanticKITTI (Table \ref{mainResults0}) and +10.84 mIoU/+17.62 IoU on SemanticTartanAir (Table \ref{mainResults2}). On SemanticKITTI, our OccDepth achieves much better results than MonoScene$^{st}$. Furthermore, compared with MonoScene$^{st}$, our OccDepth also achieves considerable improvements ([+2.33 mIoU, +4.26 IoU]). This implies that OccDepth can give more elaborate geometric structure than MonoScene$^{st}$. On SemanticTartanAir, OccDepth achieves the best SSC mIoU with improvements (+2.17 mIoU) than MonoScene$^{st}$. However, due to that 2.5D/3D methods have natural advantages on geometry, the SC IoU of OccDepth does not achieve the best. 
	
	\textbf{Comparison with 2.5D/3D methods.} We also compare OccDepth with the original 2.5D/3D input baselines. The results are listed in Table \ref{Baselines}. In SemanticKITTI (hidden test set), though OccDepth only uses stereo images that has a much less horizontal field of view~(FOV) than lidar ($~82^\circ$ vs. $180^\circ$), OccDepth achieves comparable results with 2.5D/3D-input methods. This result demonstrates that OccDepth has a relatively better completion ability. In NYUv2 (test set), because there are no stereo images, our OccDepth generates virtual stereo images through RGB image and depth map. our OccDepth achieves the best mIoU and IoU ([+0.8 IoU, +1.7 mIoU]) than 2.5D/3D methods. 
	In SemanticTartanAir (test set), we use GT depth as the input for these 2.5D/3D methods here, so the accuracy is very high. On the other hand, our OccDepth has a close mIoU result compared with 2.5D/3D-input methods while we do not use the GT depth. And compared with RGB-inferred method, our OccDepth has a much higher IoU and mIoU ([+17.6 IoU, +10.9 mIoU]).

	\subsubsection{Qualitative Results}
	\label{QualitativeResults}
 	\begin{figure*}[!t]
            \centering
		\begin{subfigure}{\linewidth}
			\centering
			\footnotesize
			\renewcommand{\arraystretch}{0.0}
			\setlength{\tabcolsep}{0.003\textwidth}
			\newcolumntype{P}[1]{>{\centering\arraybackslash}m{#1}}
			\captionsetup{font=scriptsize}
			\begin{tabular}{P{0.11\textwidth} P{0.13\textwidth} P{0.13\textwidth} P{0.13\textwidth} P{0.13\textwidth} P{0.13\textwidth} P{0.13\textwidth}}
				Input & AICNet$^\text{st}$~ & LMSCNet$^\text{st}$~  & JS3CNet$^\text{st}$~  & MonoScene & OccDepth \scriptsize{(ours)} & Ground Truth
				\\
				\includegraphics[width=\linewidth]{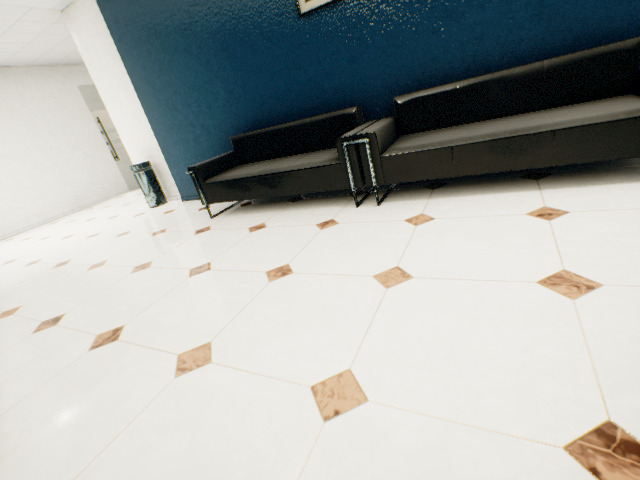} &
				\includegraphics[width=\linewidth]{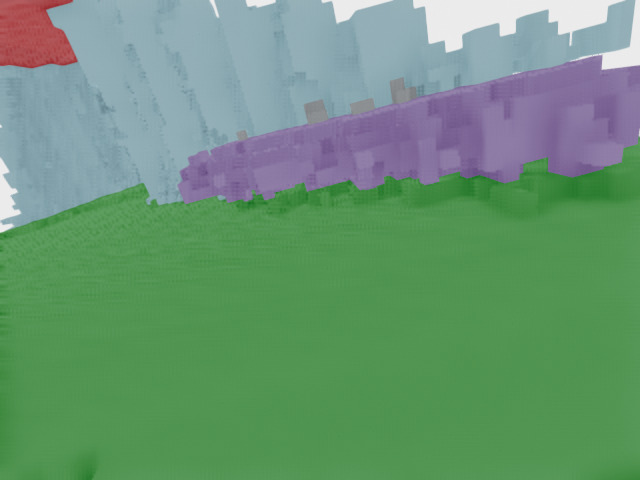} &
				\includegraphics[width=\linewidth]{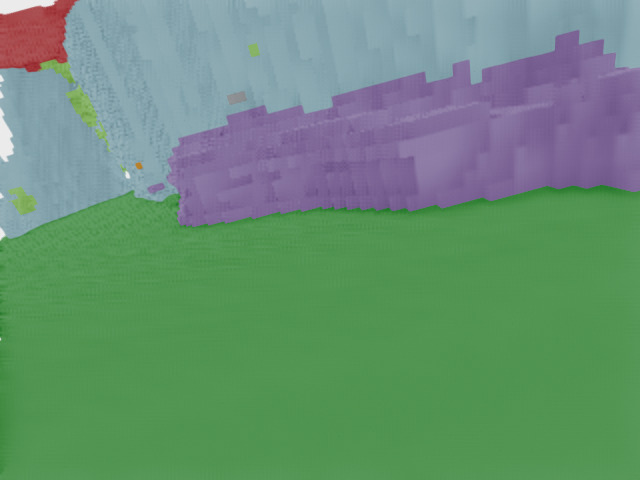} &
				\includegraphics[width=\linewidth]{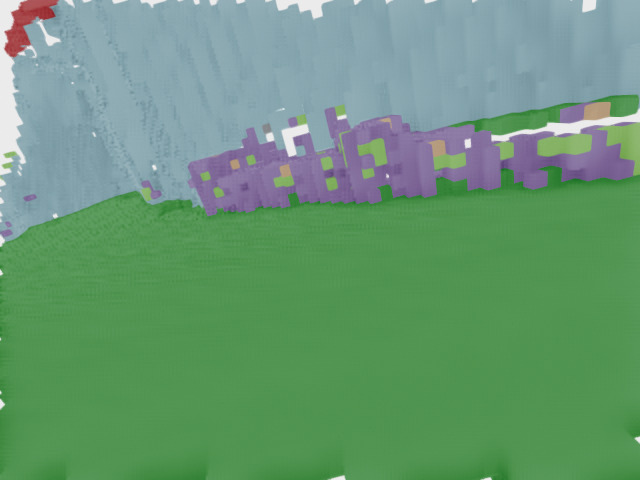} &
				\includegraphics[width=\linewidth]{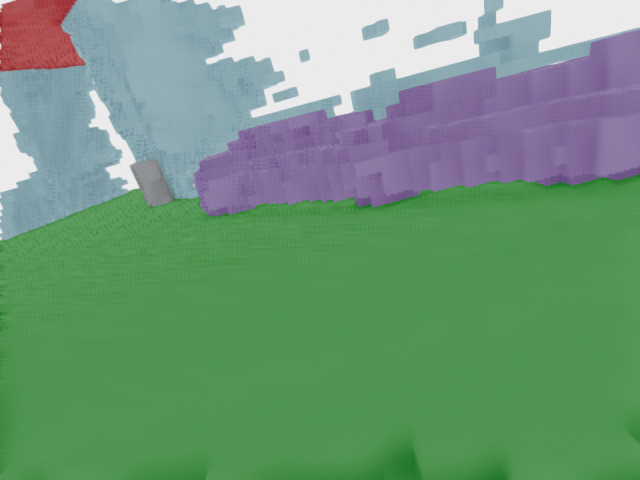} &
				\includegraphics[width=\linewidth]{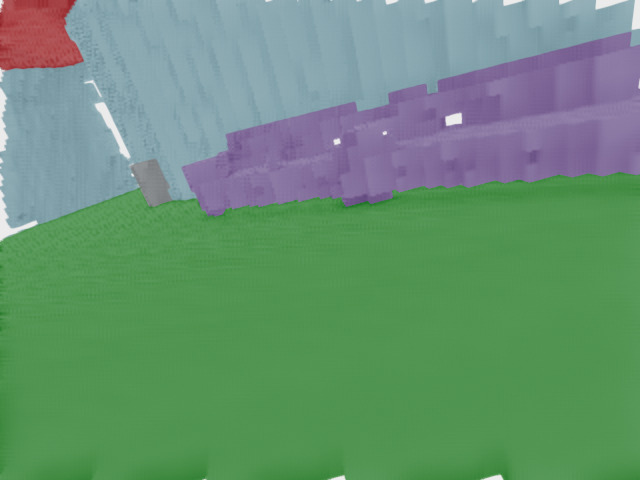} &
				\includegraphics[width=\linewidth]{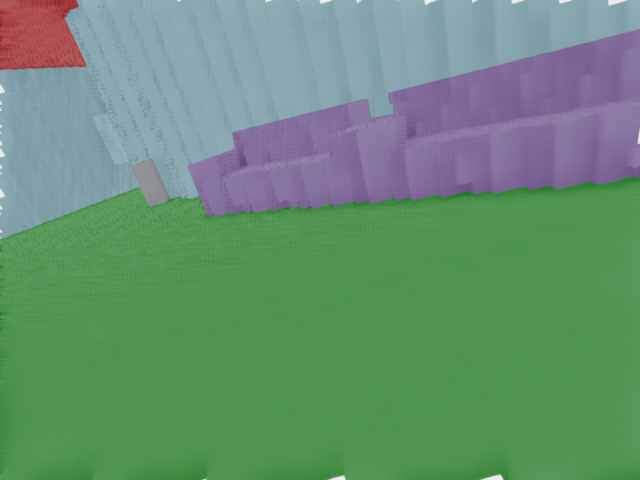}
				\\
				\includegraphics[width=\linewidth]{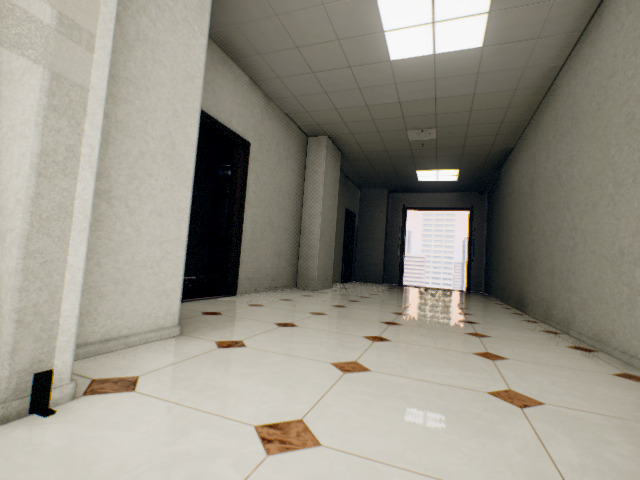} &
				\includegraphics[width=\linewidth]{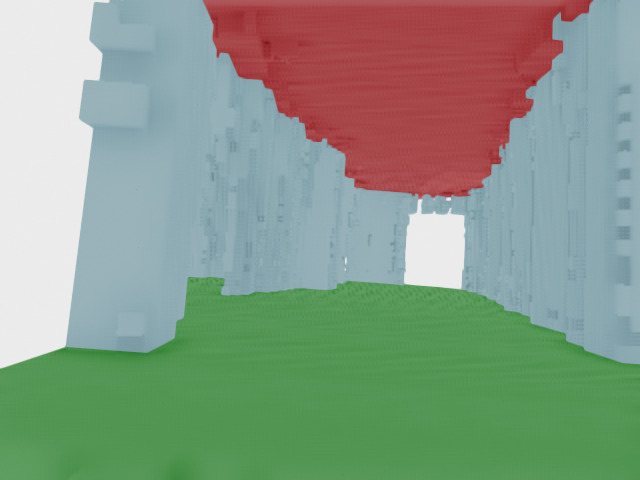} &
				\includegraphics[width=\linewidth]{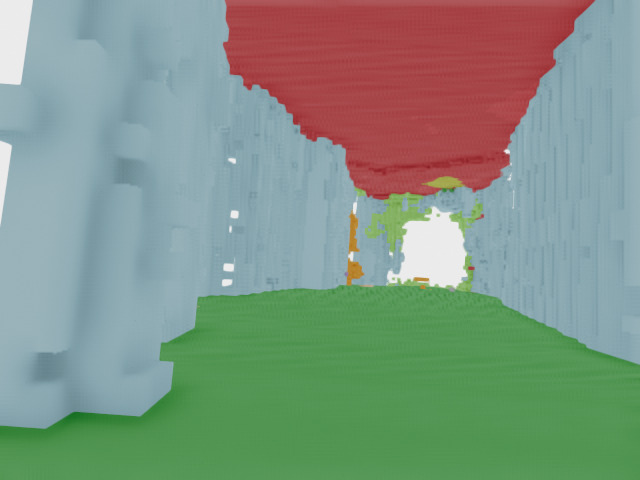} &
				\includegraphics[width=\linewidth]{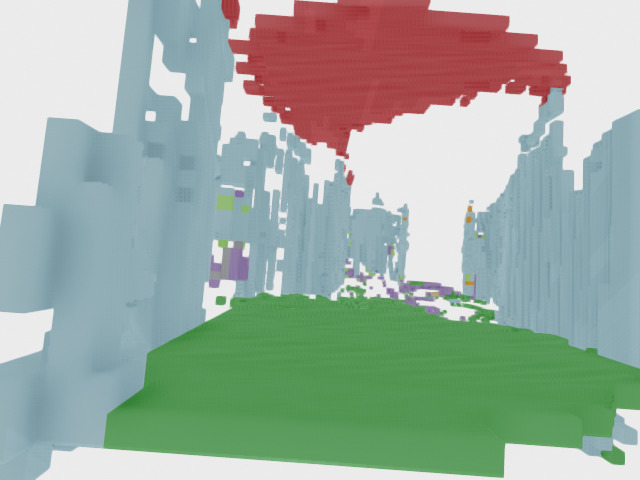} &
				\includegraphics[width=\linewidth]{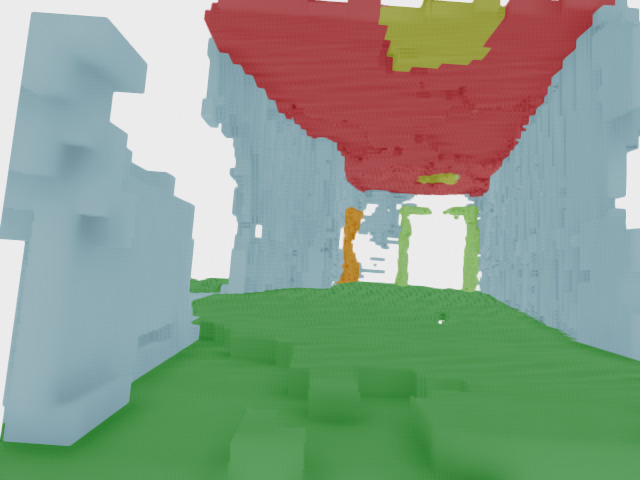} &
				\includegraphics[width=\linewidth]{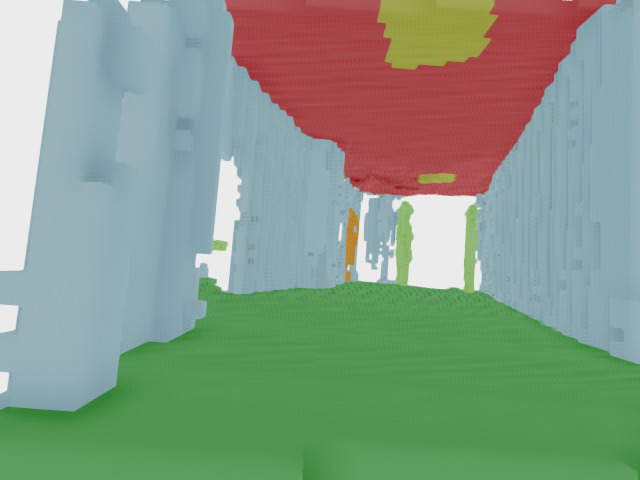} &
				\includegraphics[width=\linewidth]{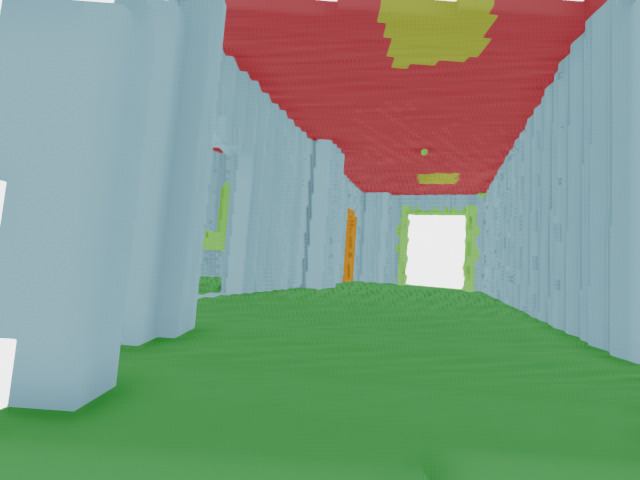}
				\\
				\includegraphics[width=\linewidth]{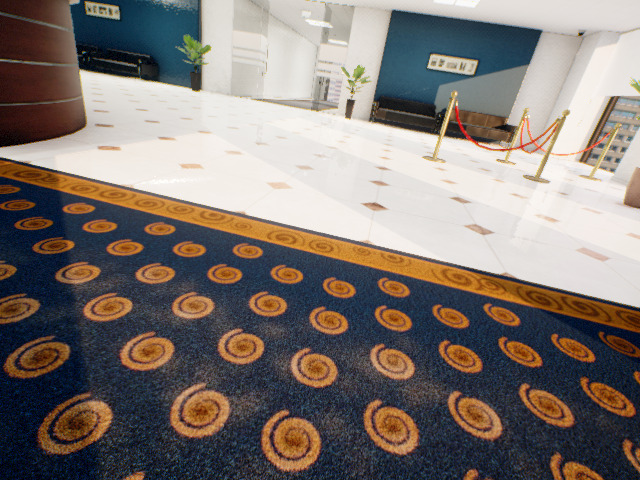} &
				\includegraphics[width=\linewidth]{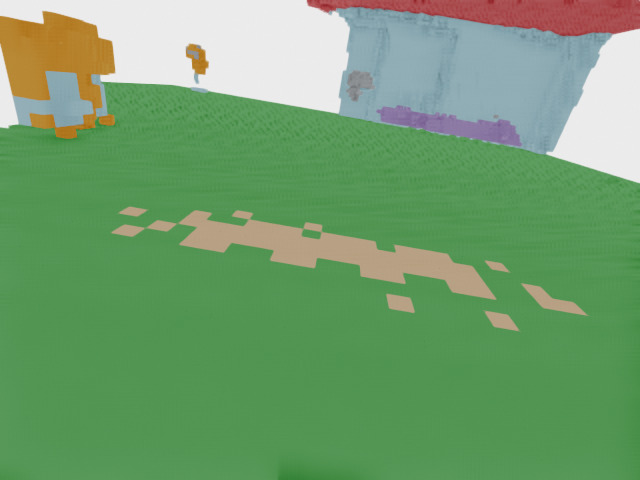} &
				\includegraphics[width=\linewidth]{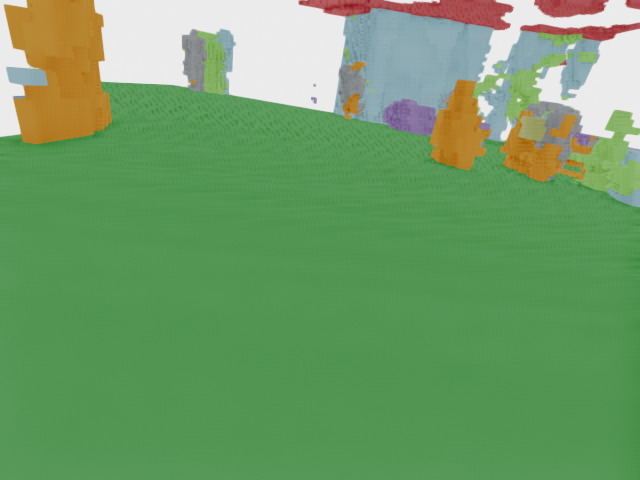} &
				\includegraphics[width=\linewidth]{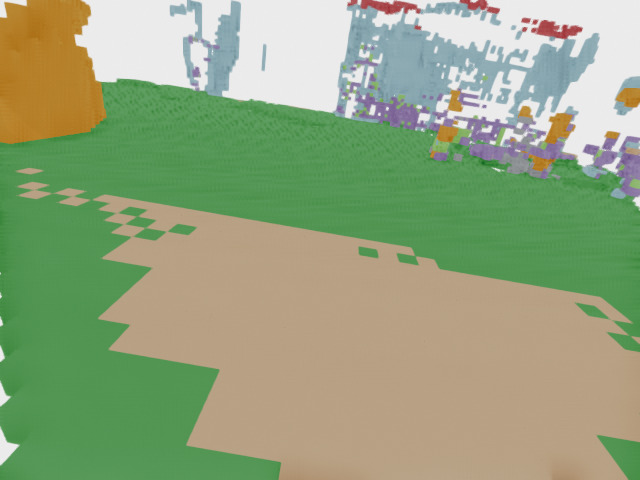} &
				\includegraphics[width=\linewidth]{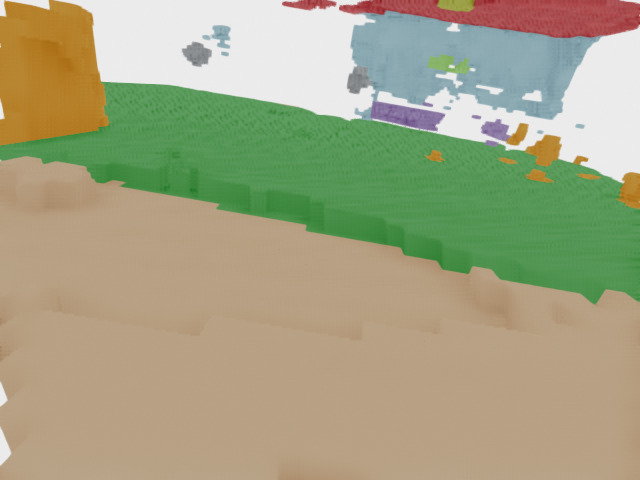} &
				\includegraphics[width=\linewidth]{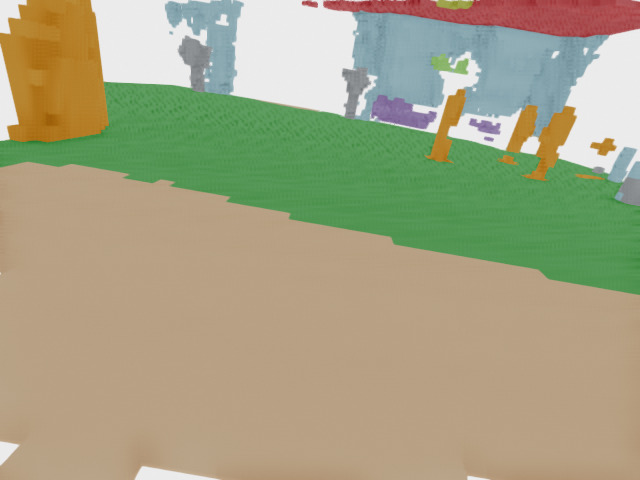} &
				\includegraphics[width=\linewidth]{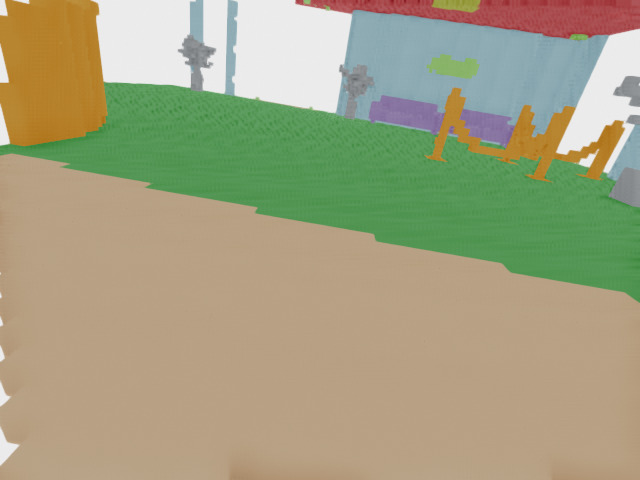}
			\end{tabular}
			\caption{\textbf{SemanticTartanAir (test set).}\\}
			\label{fig:qualitative_TartanAir}
		\end{subfigure}
		\begin{subfigure}{\linewidth}
			\centering
			\newcolumntype{P}[1]{>{\centering\arraybackslash}m{#1}}
			\setlength{\tabcolsep}{0.001\textwidth}
			\renewcommand{\arraystretch}{0}
			\footnotesize
			\captionsetup{font=scriptsize}
			\begin{tabular}{P{0.14\textwidth} P{0.14\textwidth} P{0.14\textwidth} P{0.14\textwidth} P{0.14\textwidth} P{0.14\textwidth} P{0.14\textwidth}}
				Input & AICNet$^\text{st}$~ & LMSCNet$^\text{st}$~  & JS3CNet$^\text{st}$~  & MonoScene & OccDepth \scriptsize{(ours)} & Ground Truth
				\\
				\includegraphics[width=\linewidth]{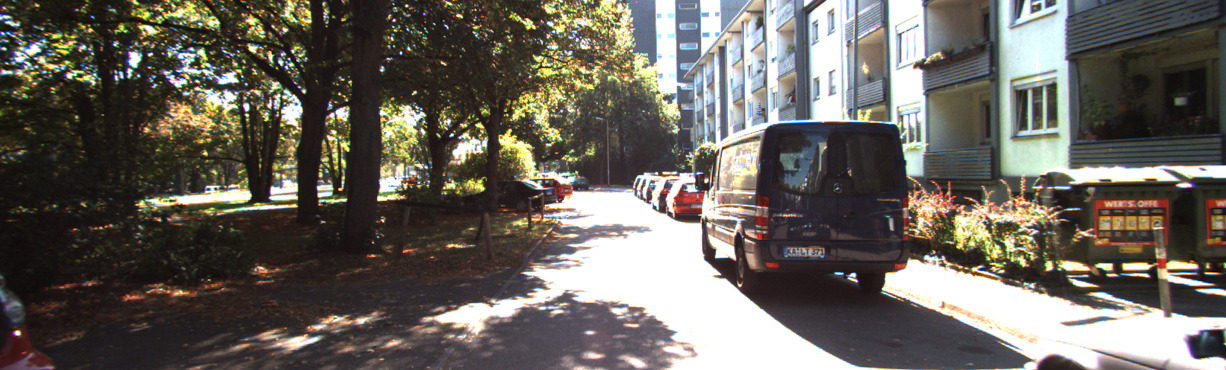} &
				\includegraphics[width=\linewidth]{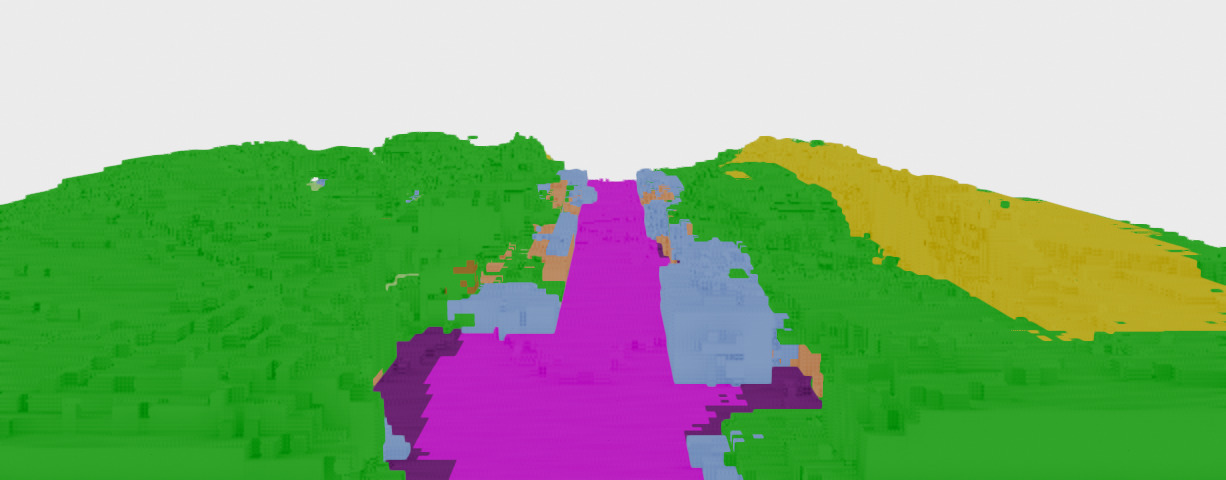} &
				\includegraphics[width=\linewidth]{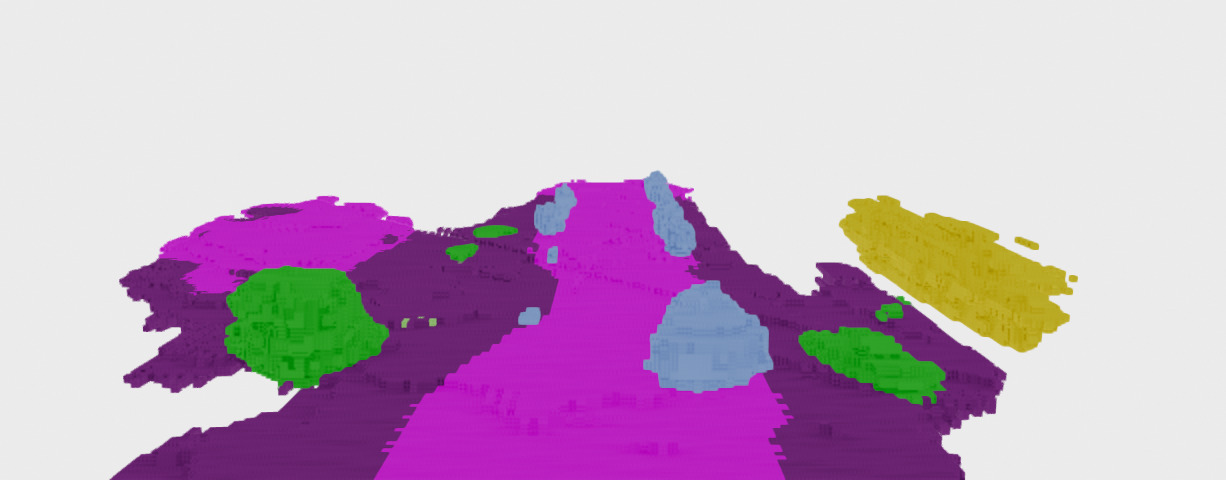} &
				\includegraphics[width=\linewidth]{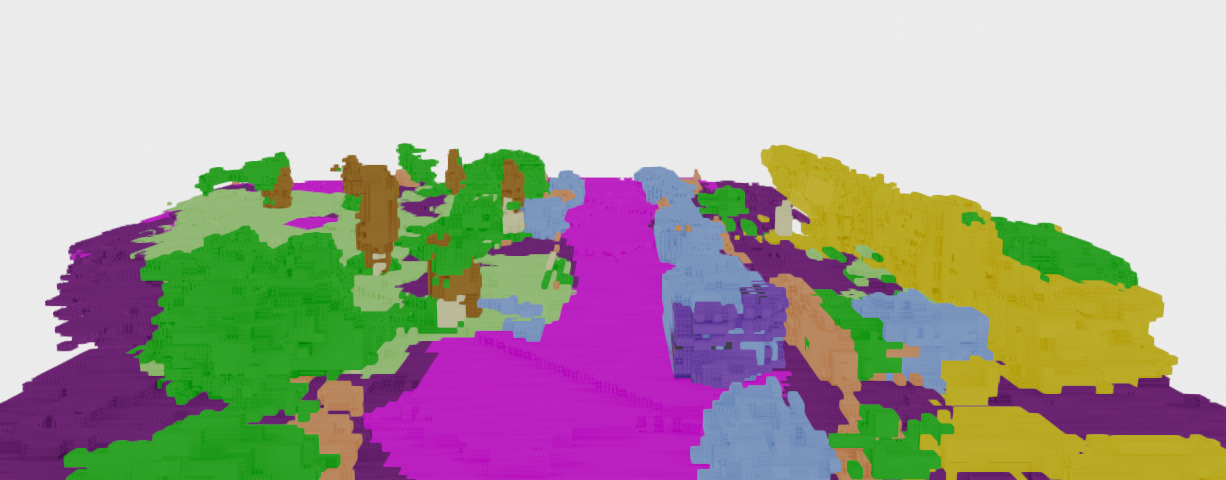} &
				\includegraphics[width=\linewidth]{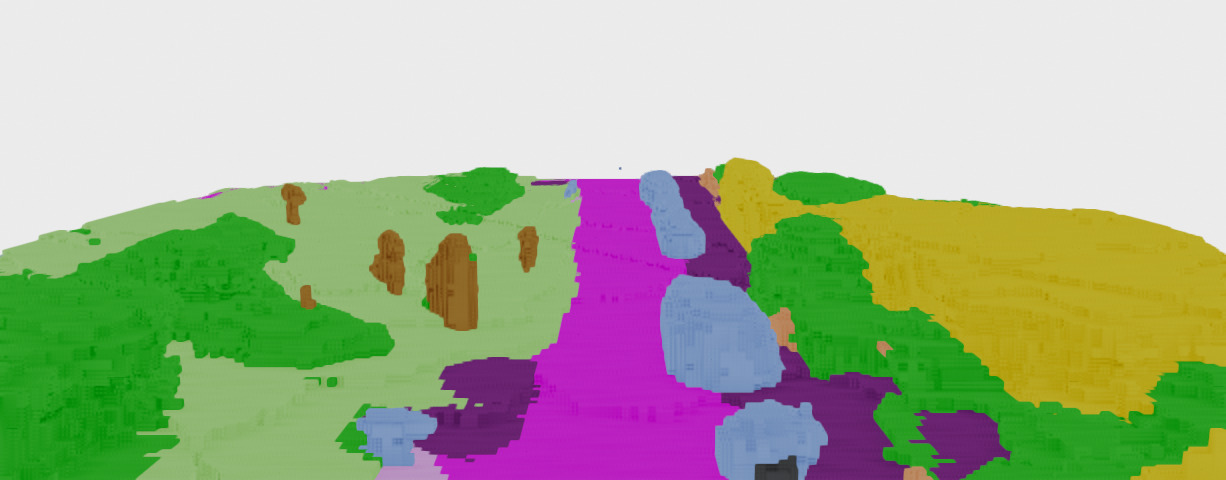} &
				\includegraphics[width=\linewidth]{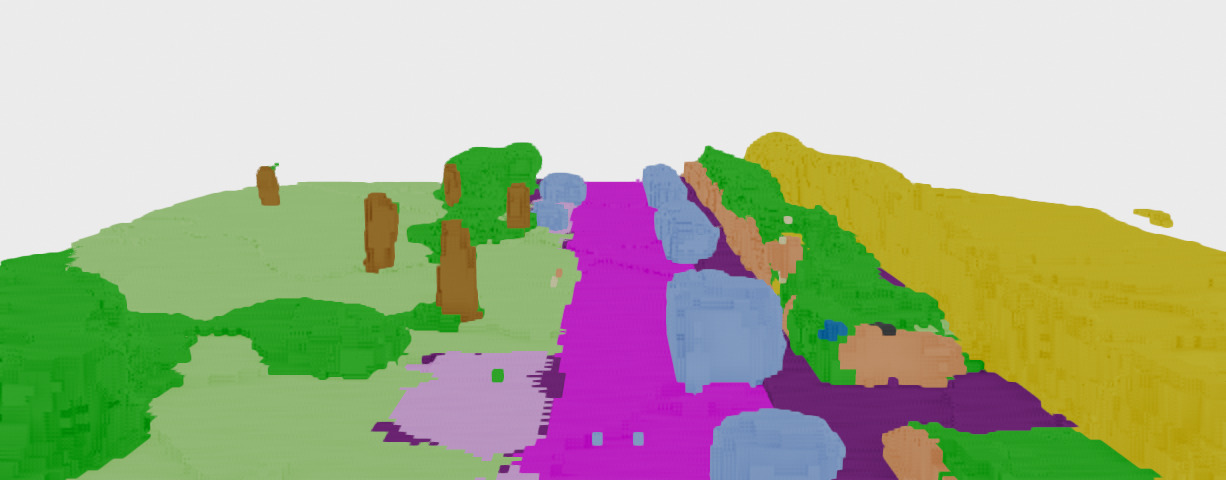} &
				\includegraphics[width=\linewidth]{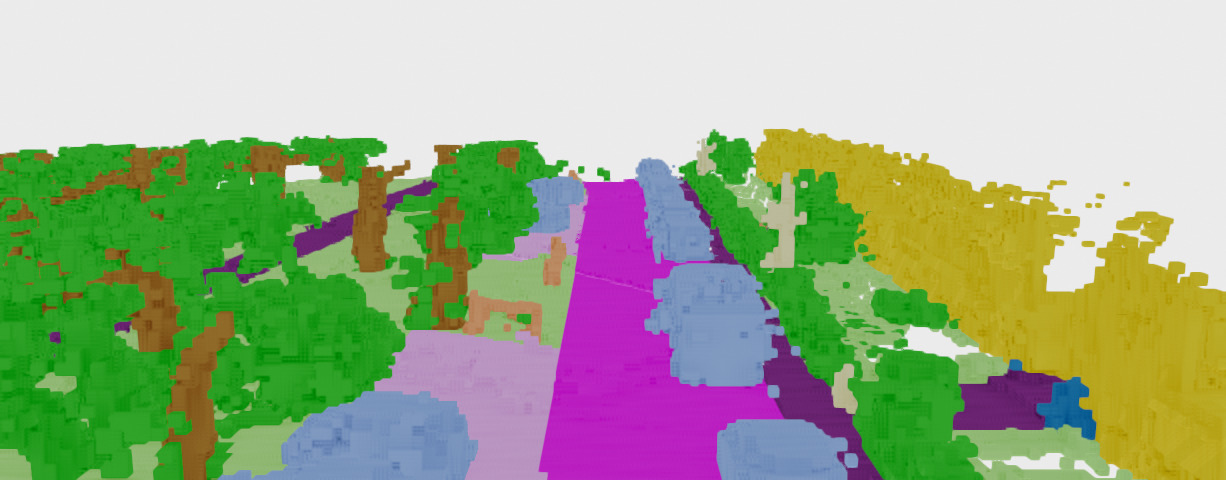}
				\\
				\includegraphics[width=\linewidth]{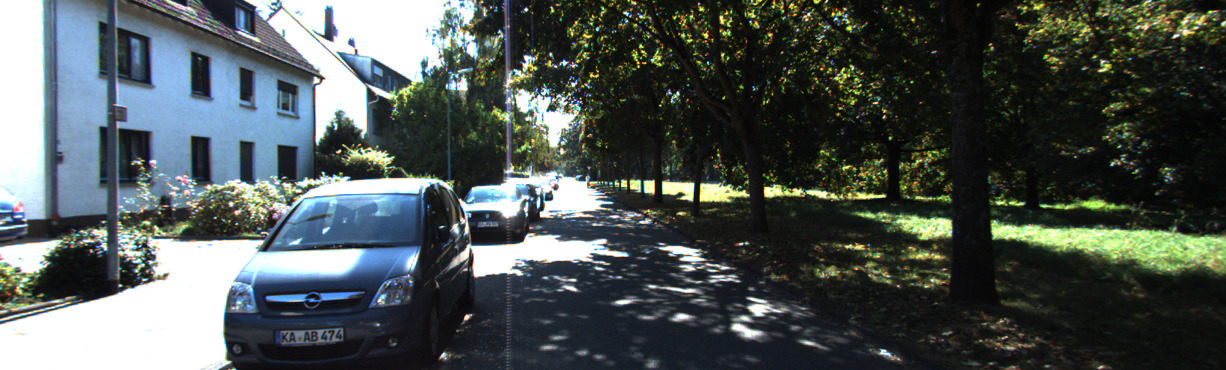} &
				\includegraphics[width=\linewidth]{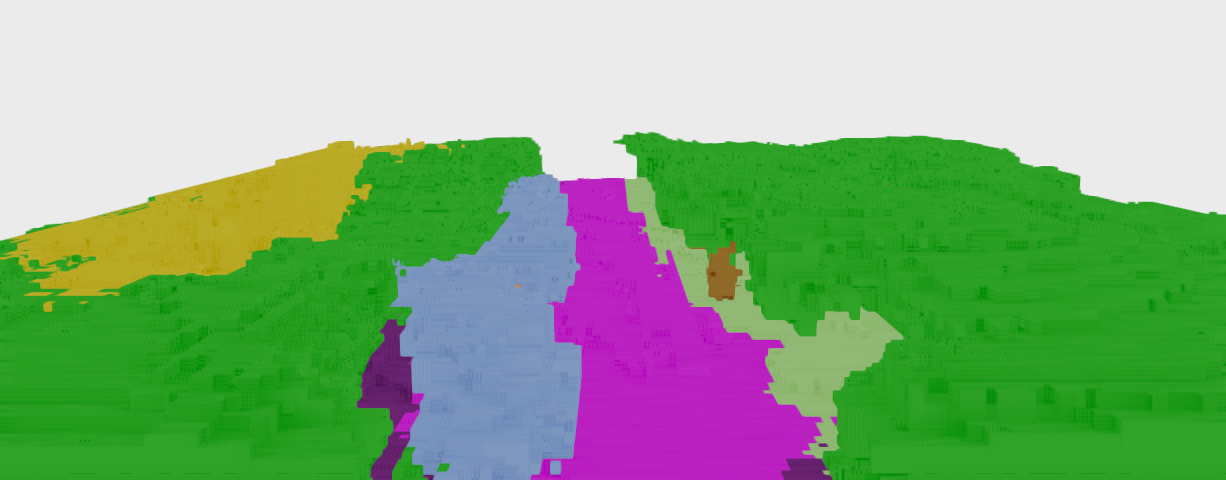} &
				\includegraphics[width=\linewidth]{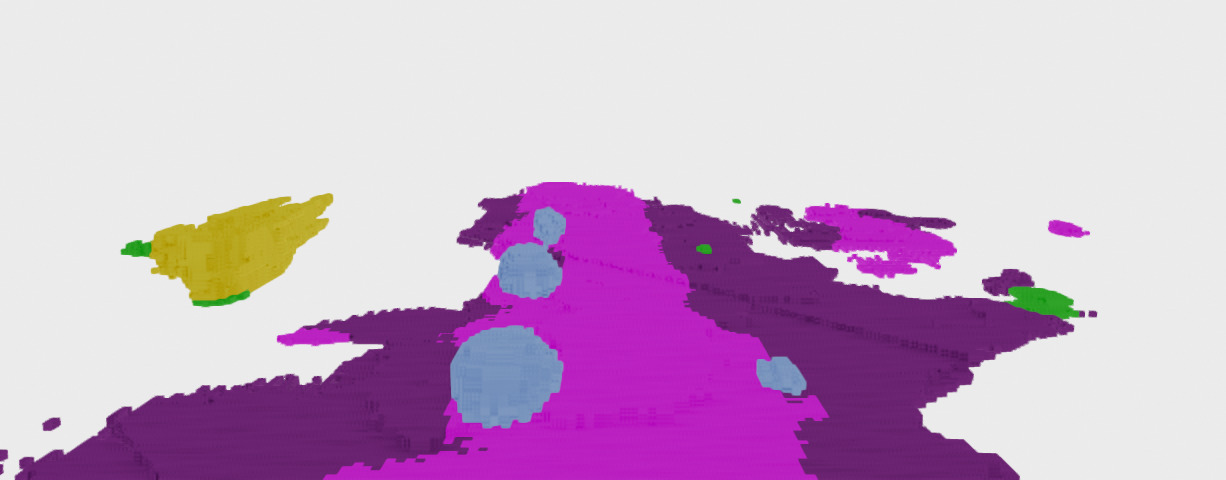} &
				\includegraphics[width=\linewidth]{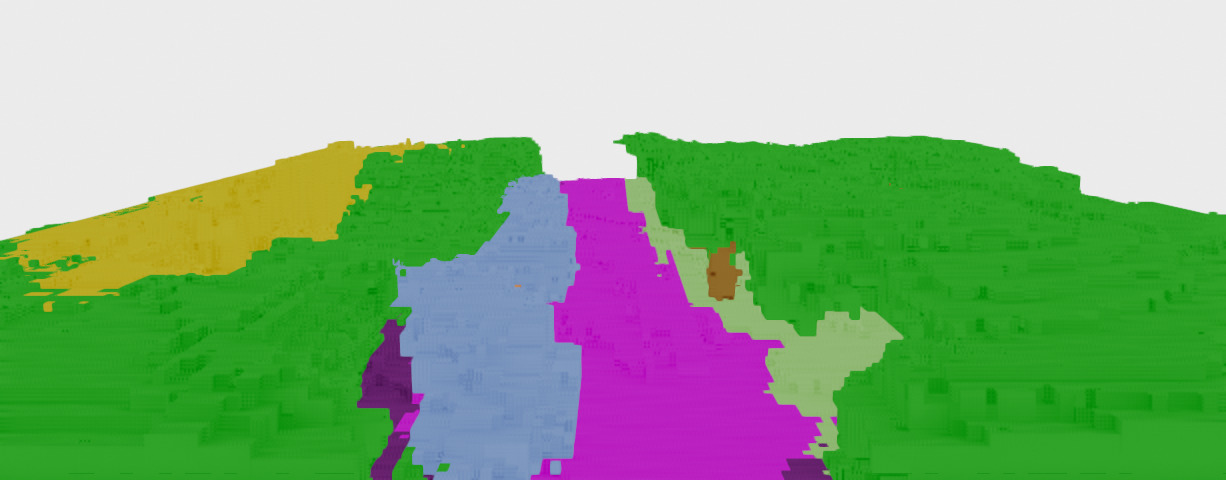} &
				\includegraphics[width=\linewidth]{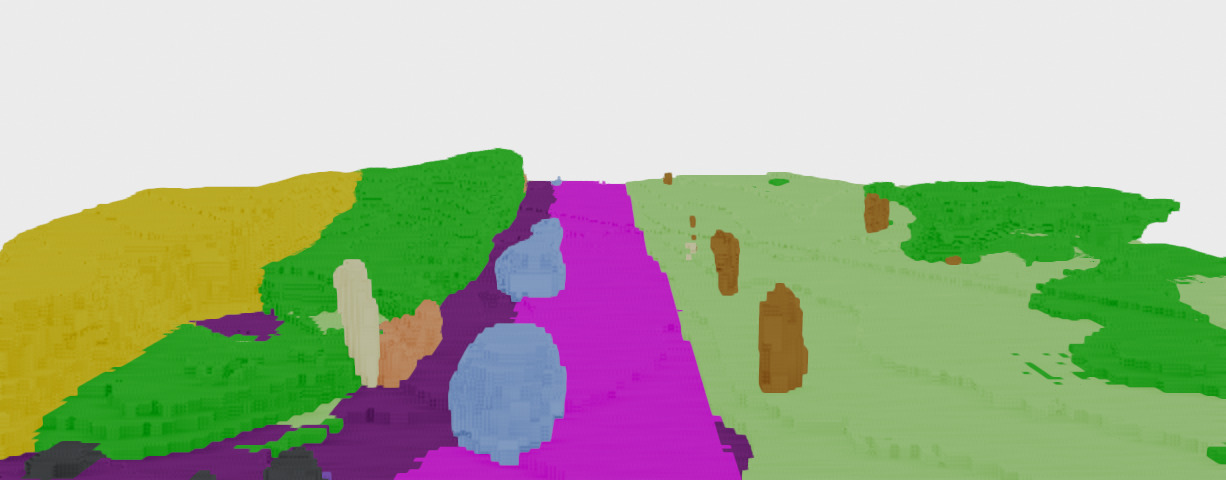} &
				\includegraphics[width=\linewidth]{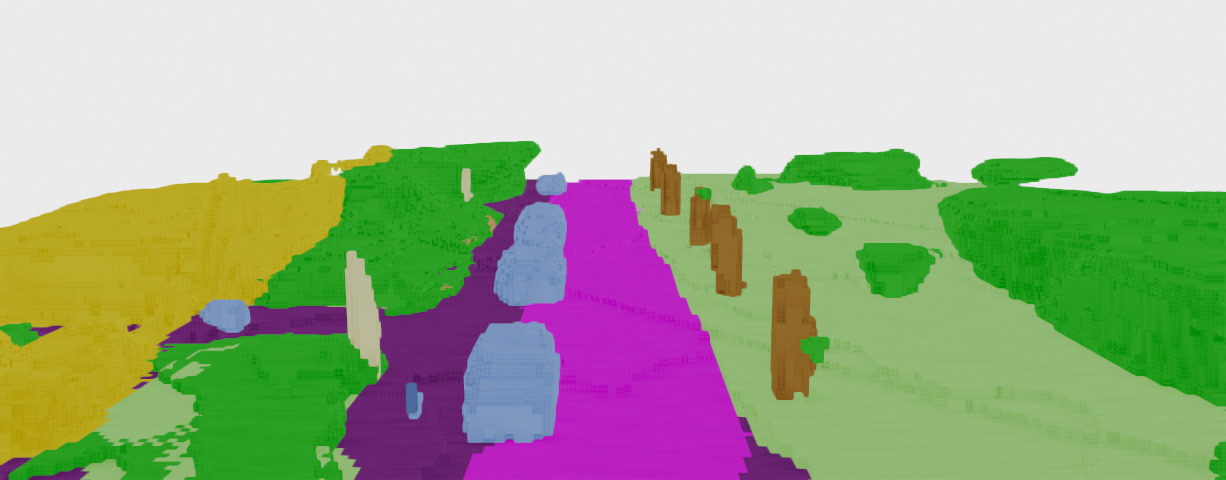} &
				\includegraphics[width=\linewidth]{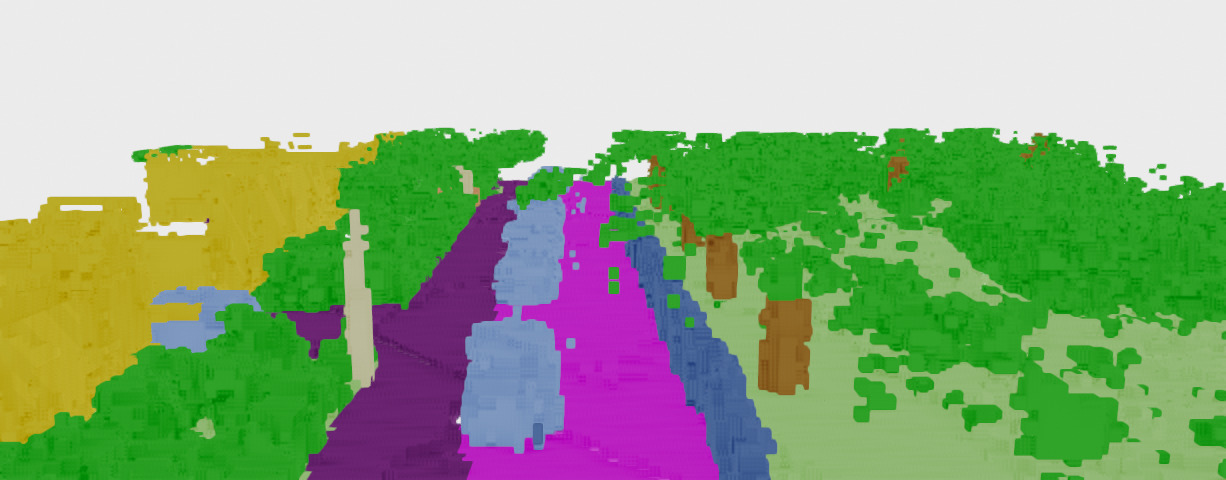}
				\\
				\includegraphics[width=\linewidth]{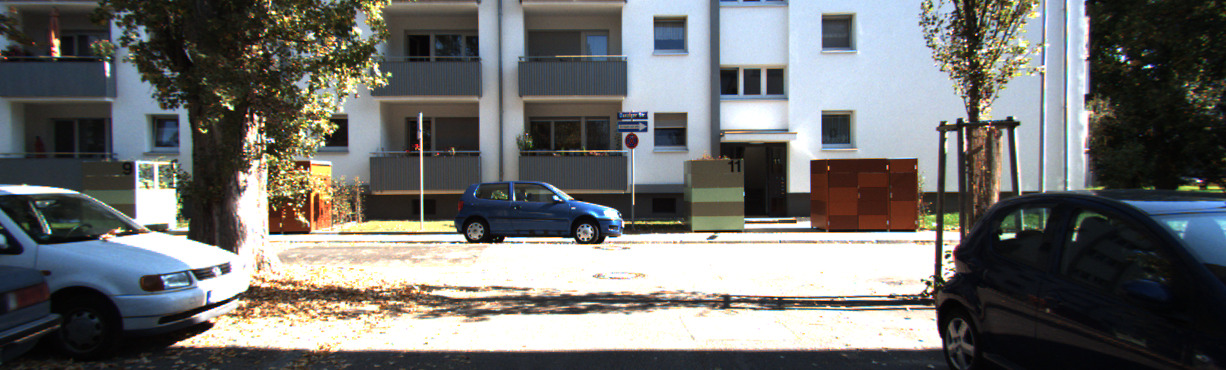} &
				\includegraphics[width=\linewidth]{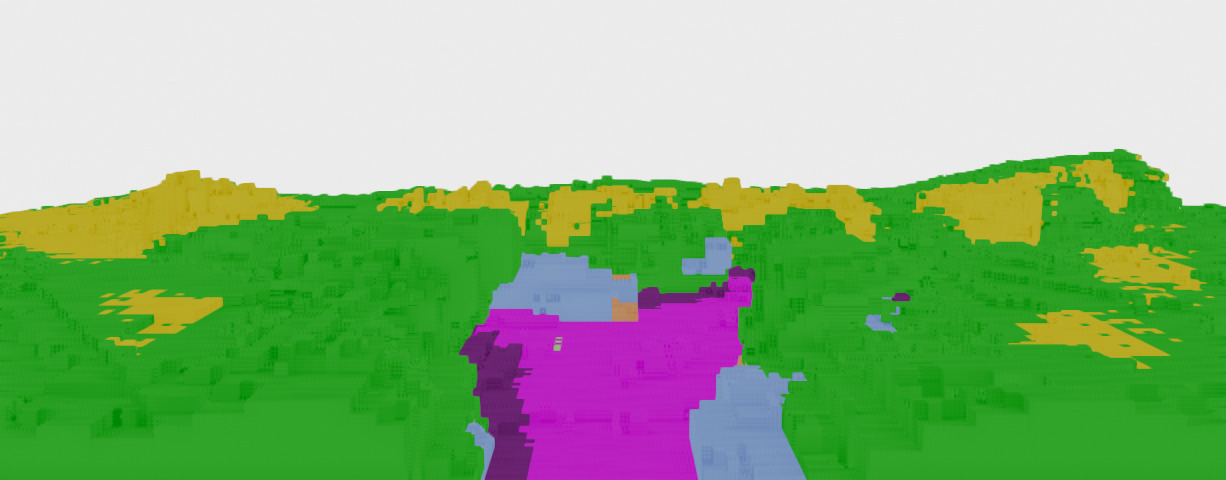} &
				\includegraphics[width=\linewidth]{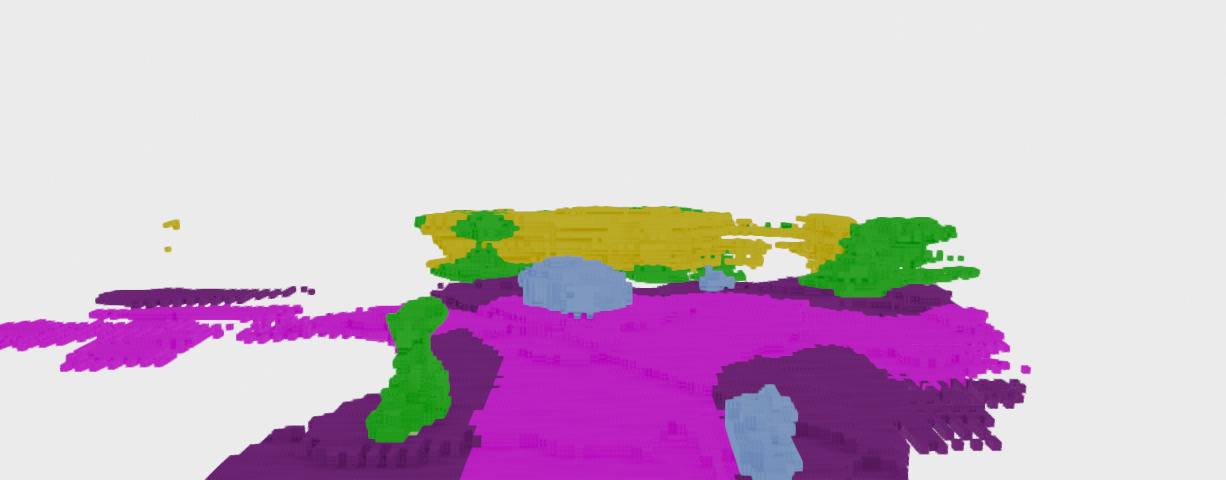} &
				\includegraphics[width=\linewidth]{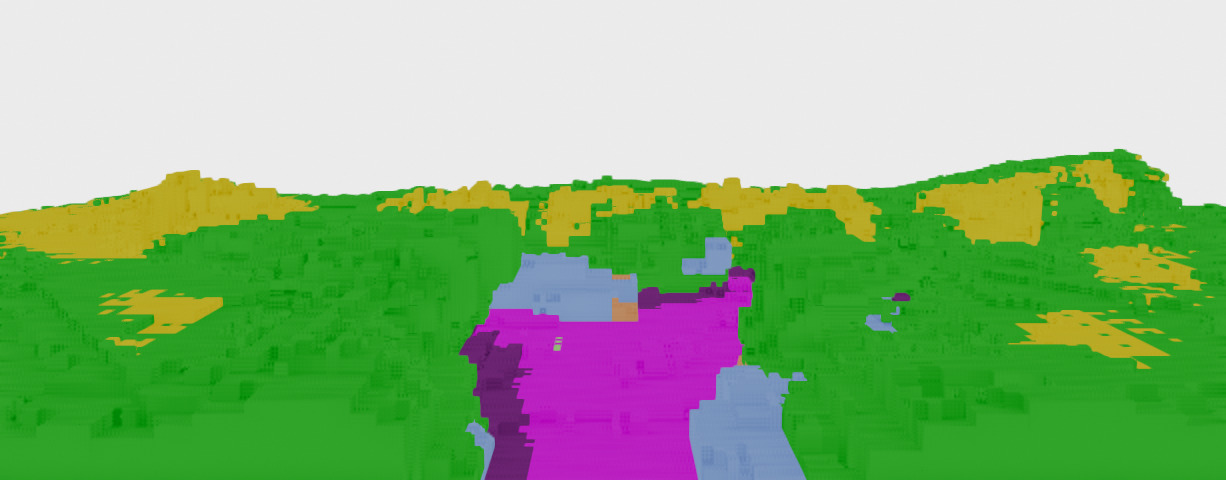} &
				\includegraphics[width=\linewidth]{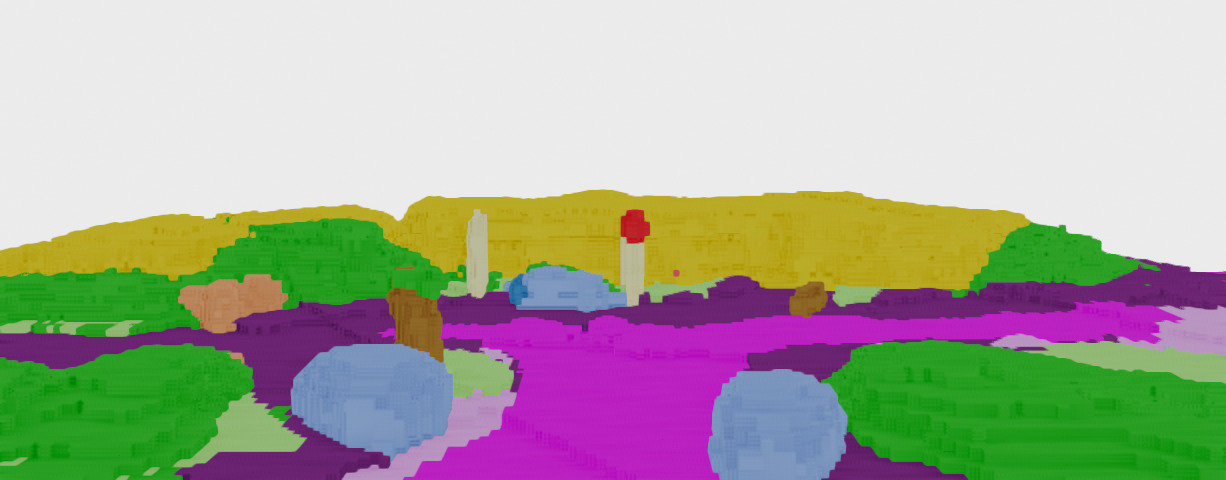} &
				\includegraphics[width=\linewidth]{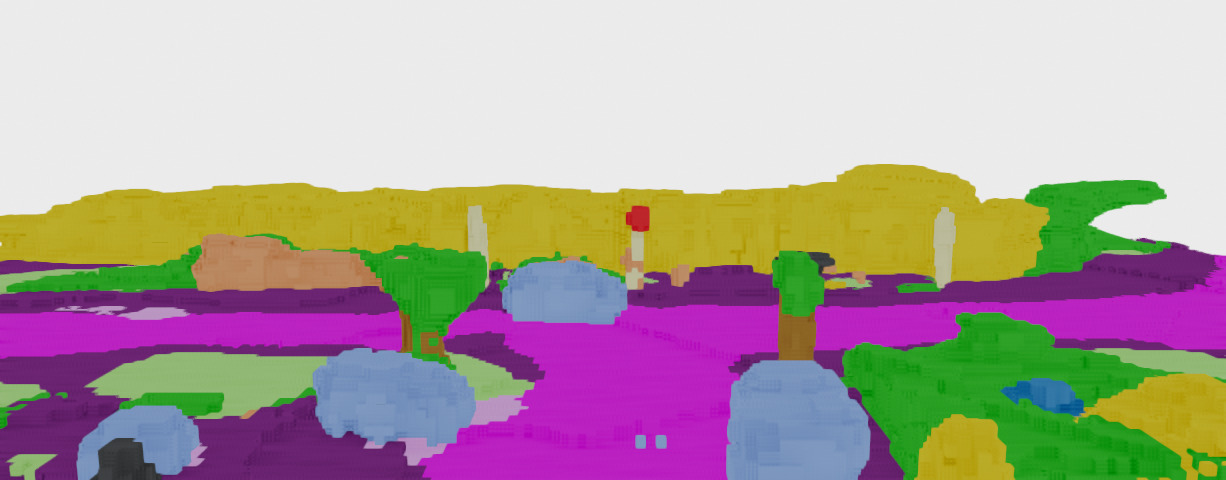} &
				\includegraphics[width=\linewidth]{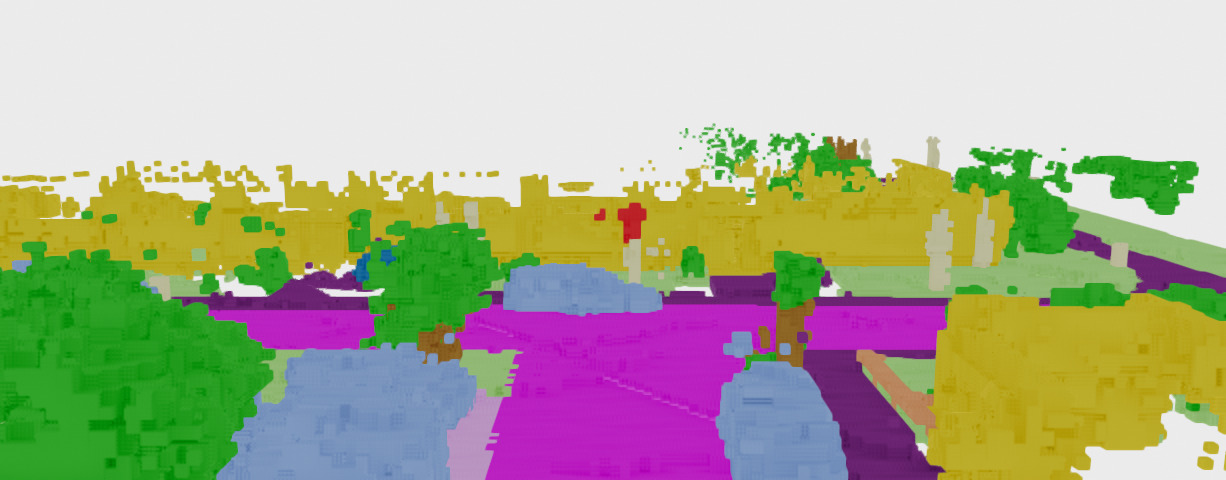}
			\end{tabular}
			\caption{\textbf{SemanticKITTI~ (val set).}}
			\label{fig:qualitative_kitti}
		\end{subfigure}
		\caption{Qualitative study on (\subref{fig:qualitative_TartanAir}) SemanticTartanAir and (\subref{fig:qualitative_kitti}) SemanticKITTI.
			The input is shown on the leftmost and the ground truth is shown on the rightmost.
			OccDepth captures better scene layout on both datasets.}
		\label{fig:qualitative}
	\end{figure*}
	We compare our SSC outputs with the above mentioned methods in Figure \ref{fig:qualitative}.
	To show the advantages of our method in visualization, the input image (leftmost column) and  its corresponding ground truth (rightmost) are also listed.

	On indoor scenes (SemanticTartanAir, Figure \ref{fig:qualitative_TartanAir}), although all methods correctly capture global scene layouts, OccDepth has better restoration effect of object edges, such as sofa (row~1) and ceiling lights (row~2) and carpet (row~3).

	On outdoor scenes (SemanticKITTI, Figure \ref{fig:qualitative_kitti}), compared to baseline methods, the spatial and semantic prediction results of OccDepth are obviously better.
	For example, accurate identification of road signs (row~1), trunk (row~2), vehicle (row~2) and road (row~3) can be achieved by OccDepth.

	\subsection{Ablation Study}
	\label{AblationStudy}

	\subsubsection{Architectural Components}
	Table \ref{ArchitectureAblation} shows that all architectural components contribute to the best results.	
	The MonoScene experiment shows the accuracy of single-view baseline. For ``Ours w/o Stereo-SFA", the multi-view fusion process is conducted by a naive average operator.
	When we use Stereo-SFA (Sec. \ref{StereoFeatureAssignment}) for multi-view fusion, the accuracy has been significantly improved ([+1.98 IoU, +0.52 mIoU]).
	The OAD (Sec. \ref{sectionOccupancyAwareDepth}) module also contributes to accuracy improvement ([+0.06 IoU, +0.65 mIoU]).
	And the depth distillation (Sec. \ref{sectionOccupancyAwareDepth}) with stereo depth data can further improve the accuracy ([+0.02 IoU, +0.51 mIoU]).
	Finally, the best accuracy ([+0.02 IoU, +0.51 mIoU]) can be obtained with tricks of the mitigating over-fitting  (Sec. \ref{sectionTMO}).

	Though Stereo-SFA module brings significant improvements on both IoU and mIoU, it also increases the amount of computation by 42\%. In contrast, the OAD module only brings significant improvement on mIoU with almost no increase in computation.

	\begin{table}
		\centering
            \resizebox{0.49\textwidth}{!}{
		\begin{tabular}{l|cc}
			\toprule
			Method & IoU $\uparrow$& mIoU $\uparrow$\\
			\midrule
			Ours & \textbf{45.25} &  \textbf{16.20} \\
			Ours w/o TMO & 45.05 & 15.80 \\
			Ours w/o Depth Distilation & 45.03 & 15.29 \\
			Ours w/o OAD & 44.97 & 14.63 \\
			Ours w/o Stereo-SFA &  42.99 &  14.11 \\
			MonoScene \cite{Cao_2022_CVPR} & 37.12 & 11.50 \\
			\bottomrule
		\end{tabular}}
		\caption{Architecture ablation. Results are reported on SemanticKITTI \textit{val}.}
		\label{ArchitectureAblation}
	\end{table}

 \begin{table*}
	\centering
	\begin{subtable}[t]{0.3\linewidth}
            \centering
		\resizebox{0.8\textwidth}{!}{
		\begin{tabular}{l|cc}
			\toprule
			Fusion methods & IoU $\uparrow$& mIoU $\uparrow$\\
			\midrule
			Mean&  39.64&  12.39 \\
			Concat & 41.30 & 13.10  \\
			Stereo-SFA& \textbf{42.60} & \textbf{13.15} \\
			\bottomrule
		\end{tabular}}
		\caption{Stereo-SFA module.}
		\label{StudyofSFA}
	\end{subtable}
	\begin{subtable}[t]{0.34\linewidth}
		\centering
		\resizebox{0.95\textwidth}{!}{
		\begin{tabular}{l|cc}
		\toprule
		Depth Discretization Method & IoU $\uparrow$& mIoU $\uparrow$\\
		\midrule
		UD & 35.36 & 10.00 \\
		SID \cite{fu2018deep} & 35.30 & 10.15 \\
		LID \cite{tang2021center3d} & \textbf{35.61} & \textbf{10.18}\\
		\bottomrule
	\end{tabular}}
	\caption{Depth discretization methods. }
	\label{StudyofOADDepthDiscre}
	\end{subtable}
	\begin{subtable}[t]{0.35\linewidth}
		\centering
	\resizebox{0.9\textwidth}{!}{
	\begin{tabular}{l|cc}
	\toprule
	Depth Distillation Data Source & IoU $\uparrow$& mIoU $\uparrow$\\
	\midrule
	w/o Depth & 45.03 & 15.30 \\
	Lidar Depth & 43.86 & 15.19 \\
	Stereo Depth & \textbf{45.05} & \textbf{15.80} \\
	\bottomrule
\end{tabular}}
\caption{Depth distillation data source.}
\label{StudyofOADDepthDistill}
\end{subtable}
	\caption{Ablation study of proposed modules. (a) Ablation study of Stereo-SFA module. (b) Ablation study of Depth distillation data source in OAD module. (c)Ablation study of Depth distillation data source in OAD module. w/o Depth means no depth distillation is used, Lidar depth refers to the depth image generated by lidar point cloud and Stereo Depth refers to the depth image generated by LEAStereo model. Results are reported on SemanticKITTI \textit{val}.}
\label{ablationAll}
\end{table*}

	\subsubsection{Effectiveness of Stereo-SFA module}
	We ablate our Stereo-SFA module on SemanticKITTI (validation set) to show the advantages of Stereo-SFA module. 2D U-Net based on pre-trained EfficientNet-B3 \cite{pmlrv97tan19a} is used to generate 2D features with feature channels 24. Table \ref{StudyofSFA} shows SC IoU and SSC mIoU results of different stereo fusion methods. The "Mean" fusion method means that the average value of left 2D feature and right 2D feature is assigned to corresponding voxel. We also implemented a better fusion method named "Concat" in Table \ref{StudyofSFA}. The "Concat" fusion method fuses feature through concatenating the left 2D feature and right 2D feature followed by a 1:2 downsampling. The results show Stereo-SFA is better than "Mean" fusion method with improvements of +2.96 IoU/+0.76 mIoU and is better than "Concat" fusion method with improvement of +1.30 IoU/+0.05 mIoU. Note that Stereo-SFA fusion method significantly improves the SC IoU. And the improvement on SC IoU proves that the geometric information is more effectively introduced into the system by our Stereo-SFA module.

	\subsubsection{Effectiveness of OAD module}
	We provide ablation studies of OAD module to validate our design choices.
	The results are shown in Tables \ref{StudyofOADDepthDiscre} and \ref{StudyofOADDepthDistill}.
	Experiment in Table \ref{StudyofOADDepthDiscre} shows the experimental results of several different depth discretization methods.
	The comparison experiments are conducted with single-view image input, and the small model (EfficientNet-B3) with feature channels 24 are used to improve the efficiency of comparative experiments.
	Note that the LID depth discretization method achieves the highest accuracy. The SID depth discretization method also achieves comparable mIoU with LID depth discretization method since SID also spends more depth bin resources on nearby objects.
	
	Experiment in Table \ref{StudyofOADDepthDistill} shows the experimental results of two different depth sources for depth distillation process.
	The lidar depth images are generated by a lidar point cloud projection process, so these depth images are sparse in image space. The stereo depth images are generated by the LEAStereo model, so the generated depth images are spatially dense due to the compensation Ability of Neural Networks. Our depth distillation method with lidar depth data achieves 15.19\% mIoU, and achieves 15.80\% mIoU with lidar depth data. A reasonable explanation for the results maybe the sparse property in image space leads to low accuracy.
	
	\section{Conclusion}
	In this paper, a depth-aware method for 3D SSC, which is named OccDepth, is presented.  We trained OccDepth on public datasets including SemanticKITTI (outdoor scene) and NYUv2 (indoor scene) dataset. Experiment results demonstrate that OccDepth is comparable to some 2.5D/3D-input methods on both indoor scene and outdoor scene. Especially, OccDepth outperforms current RGB-inferred baselines on all classes. Moreover, the experiment results on newly proposed indoor benchmark based on SemanticTartanAir verify the improvements of stereo input in indoor scene. As far as we know, OccDepth is the first stereo RGB-inferred method that is comparable to 2.5D/3D-input SSC methods.
	\bibliographystyle{named}
	\bibliography{ijcai23}
	
\end{document}